\documentclass{article} % For LaTeX2e
\usepackage{iclr2025_conference,times}

\usepackage{amsmath,amsfonts,bm}

\def\eqref#1{equation~\ref{#1}}
\def\1{\bm{1}}

\DeclareMathAlphabet{\mathsfit}{\encodingdefault}{\sfdefault}{m}{sl}
\SetMathAlphabet{\mathsfit}{bold}{\encodingdefault}{\sfdefault}{bx}{n}

\usepackage{hyperref}
\usepackage{url}

\usepackage{mathtools} % underbrackets
\usepackage{multirow} % table multirow
\usepackage{wrapfig} % side figure
\usepackage{booktabs} % professional-quality tables
\usepackage{subcaption} % subfigure
\usepackage{sidecap}

\newcommand{\resultmean}{162}
\newcommand{\resultmedian}{77}

\title{Learning Transformer-based World Models \\ with Contrastive Predictive Coding}

\author{Maxime Burchi, Radu Timofte \\
Computer Vision Lab, CAIDAS \& IFI, University of Würzburg, Germany \\
\texttt{maxime.burchi@uni-wuerzburg.de}
}

\iclrfinalcopy
\begin{document}

\maketitle

\begin{abstract}

The DreamerV3 algorithm recently obtained remarkable performance across diverse environment domains by learning an accurate world model based on Recurrent Neural Networks (RNNs). Following the success of model-based reinforcement learning algorithms and the rapid adoption of the Transformer architecture for its superior training efficiency and favorable scaling properties, recent works such as STORM have proposed replacing RNN-based world models with Transformer-based world models using masked self-attention. However, despite the improved training efficiency of these methods, their impact on performance remains limited compared to the Dreamer algorithm, struggling to learn competitive Transformer-based world models. In this work, we show that the next state prediction objective adopted in previous approaches is insufficient to fully exploit the representation capabilities of Transformers. We propose to extend world model predictions to longer time horizons by introducing TWISTER (Transformer-based World model wIth contraSTivE Representations), a world model using action-conditioned Contrastive Predictive Coding to learn high-level temporal feature representations and improve the agent performance. TWISTER achieves a human-normalized mean score of \resultmean\% on the Atari 100k benchmark, setting a new record among state-of-the-art methods that do not employ look-ahead search. We release our code at \texttt{https://github.com/burchim/TWISTER}.

\end{abstract}

\section{Introduction}

\begin{wrapfigure}{r}{0.4\textwidth}
    \centering
    \vspace{-2.5ex}
    \includegraphics[width=\linewidth]{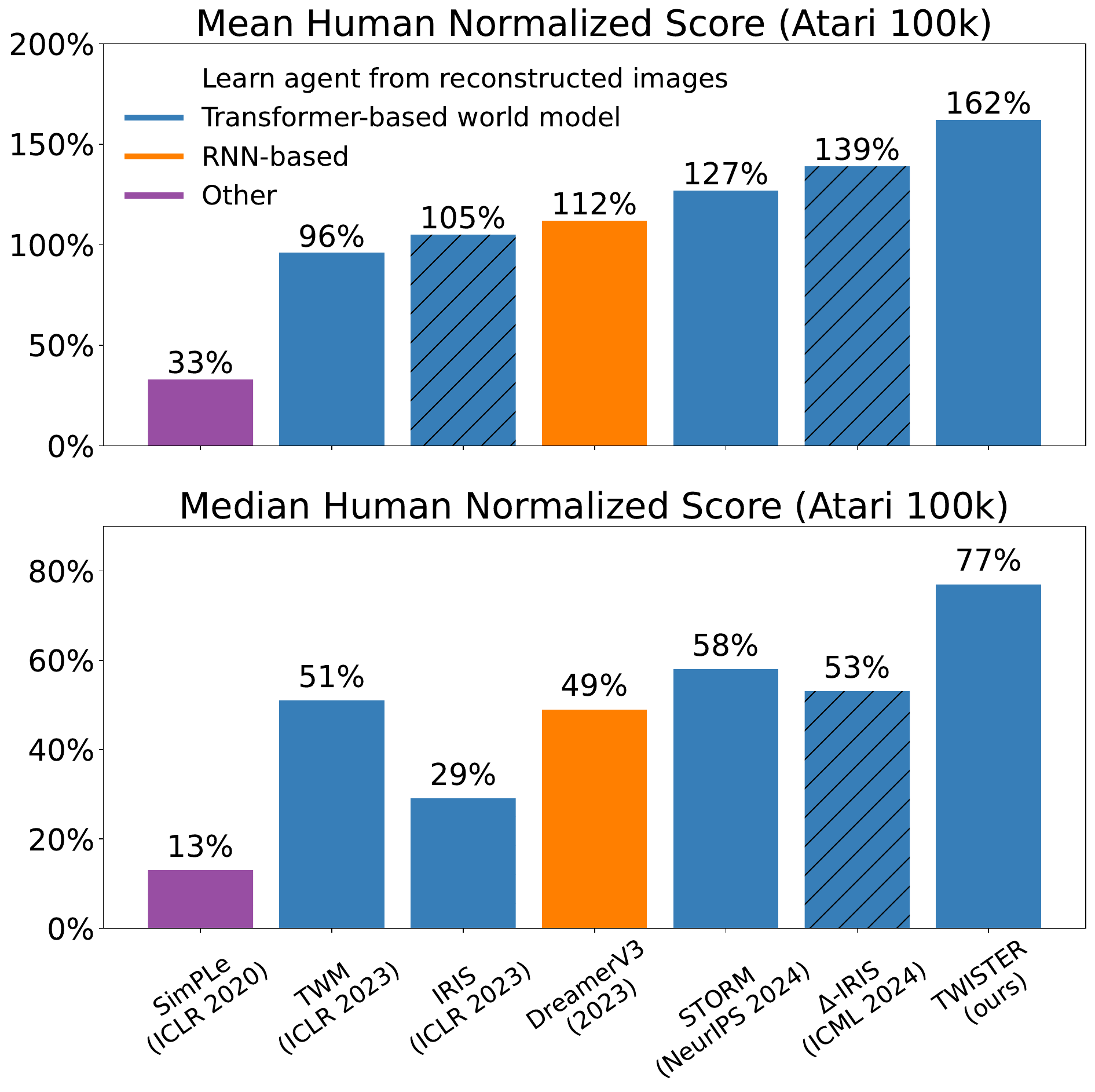}
    \vspace{-3ex} 
    \caption{Human-normalized mean and median scores of recently published model-based methods on the Atari 100k benchmark. TWISTER outperforms other model-based approaches. TWM, IRIS, STORM and $\Delta$-IRIS employ a Transformer-based world model while DreamerV3 uses a RNN-based model.}
    \vspace{-5ex}
    \label{figure:bar_results}
\end{wrapfigure}

Deep Reinforcement Learning (RL) algorithms have achieved notable breakthroughs in recent years. The growing computational capabilities of hardware systems have allowed researchers to make significant progress, training powerful agents from high-dimensional observations like images~\citep{mnih2013playing} or videos~\citep{hafner2019dream} using deep neural networks~\citep{lecun2015deep} as function approximations. Following the rapid adoption of Convolutional Neural Networks (CNNs)~\citep{lecun1989handwritten} in the field of Computer Vision for their efficient pattern recognition ability, neural networks were applied to visual reinforcement learning problems and achieved human to superhuman performance in challenging and visually complex domains like Atari games~\citep{mnih2015human, hessel2018rainbow}, the game of Go~\citep{silver2018general, schrittwieser2020mastering}, StarCraft II~\citep{vinyals2019grandmaster} and more recently, Minecraft~\citep{baker2022video, hafner2023mastering}. 

Following the success of neural networks in solving reinforcement learning problems, model-based approaches learning world models using gradient backpropagation were proposed to reduce the amount of necessary interaction with the environment to achieve strong results~\citep{kaiser2019model, hafner2019dream, hafner2020mastering, hafner2023mastering, schrittwieser2020mastering}. World models~\citep{sutton1991dyna, ha2018recurrent} summarize an agent’s experience into a predictive model that can be used in place of the real environment to learn complex behaviors. Having access to a model of the environment enables the agent to simulate multiple plausible trajectories in parallel, improving generalization, sample efficiency and decision-making via planning. 

\begin{wrapfigure}{r}{0.4\textwidth}
    \centering
    \vspace{-2ex}
    \includegraphics[width=\linewidth]{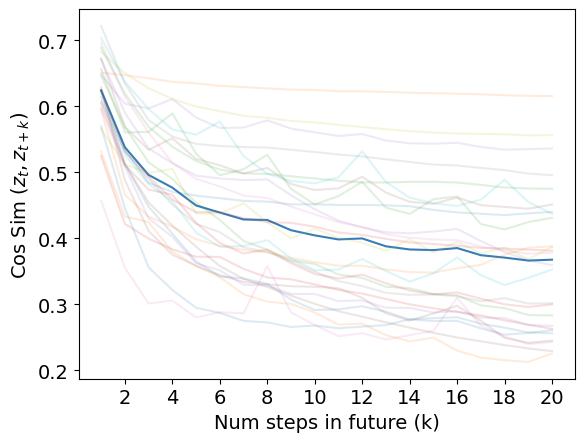}
    \caption{Cosine Similarities between TWISTER latent state $z_{t}$ and future states $z_{t+k}$ aggregated over all 26 games of the Atari 100k benchmark. We show average similarities over 5 seeds.}
    \label{figure:plot_cos_sims}
    \vspace{-2ex}
\end{wrapfigure}

Design choices for the world model have tended toward Recurrent Neural Networks (RNNs)~\citep{hafner2019learning} for their ability to model temporal relationships effectively. Following the success of the Dreamer algorithm~\citep{hafner2019dream} and the rapid adoption of the Transformer architecture~\citep{vaswani2017attention} for its superior training efficiency and favorable scaling properties compared to RNNs, research works have proposed replacing the one-layer recurrent-based world model of Dreamer with a Transformer-based world model using masked self-attention~\citep{chen2022transdreamer,micheli2022transformers,robine2023transformer}. However, despite the improved training efficiency of these methods, their impact on performance remains limited compared to the Dreamer algorithm, struggling to learn competitive Transformer-based world models. \citet{zhang2024storm} suggested that these findings may be attributed to the subtle differences between consecutive video frames. The task of predicting the next video frame in latent space may not require a complex model in contrast to other fields like Neural Language Modeling~\citep{kaplan2020scaling} where a deep understanding of the past context is essential to accurately predict the next tokens. As shown in Figure~\ref{figure:plot_cos_sims}, the cosine similarity between adjacent latent states of the world model is very high, making it relatively straightforward for the world model to predict the next state compared to more distant states. These findings motivate our work to complexify the world model objective by extending predictions to longer time horizons in order to learn higher quality feature representations and improve the agent performance.

In this work, we show that the next latent state prediction objective adopted in previous approaches is insufficient to fully exploit the representation capabilities of Transformers. We introduce TWISTER, a Transformer model-based reinforcement learning algorithm using action-conditioned Contrastive Predictive Coding (AC-CPC) to learn high-level temporal feature representations and improve the agent performance. CPC~\citep{oord2018representation} was initially applied to speech, image, and text domains as a pretraining pretext task. It also showed promising results on DeepMind Lab tasks~\citep{beattie2016deepmind} being used as an auxiliary loss for the A3C agent~\citep{mnih2016asynchronous}. Motivated by these findings, we apply the CPC objective to model-based reinforcement learning by conditioning CPC predictions on the sequence of future actions. This approach enables the world model to accurately predict the feature representations of future time steps using contrastive learning. As shown in Figure~\ref{figure:bar_results}, TWISTER sets a new record on the commonly used Atari 100k benchmark~\citep{kaiser2019model} among state-of-the-art methods that do not employ look-ahead search, achieving a human-normalized mean and median score of \resultmean\% and \resultmedian\%, respectively.
\section{Related Works}

\subsection{Model-based Reinforcement Learning}

Model-based reinforcement learning approaches use a model of the environment to simulate agent trajectories, improving generalization, sample efficiency, and decision-making via planning. Following the success of deep neural networks for learning function approximations, researchers proposed to learn world models using gradient backpropagation. While initial works concentrated on simple environments like proprioceptive tasks~\citep{silver2017predictron, henaff2017model, wang2019benchmarking, wang2019exploring} using low-dimensional observations, more recent works focus on learning world models from high-dimensional observations like images~\citep{kaiser2019model, hafner2019learning}.

One of the earliest model-based algorithms applied to image data is SimPLe~\citep{kaiser2019model}, which proposed to learn a world model for Atari games in pixel space using a convolutional autoencoder. The world model learns to predict the next frame and environment reward given previous observation frames and selected action. It is then used to train a Proximal Policy Optimization (PPO) agent~\citep{schulman2017proximal} from reconstructed images and predicted rewards. Concurrently, PlaNet~\citep{hafner2019learning} introduced a Recurrent State-Space Model (RSSM) using a Gated Recurrent Unit (GRU)~\citep{cho2014learning} to learn a world model in latent space, planning using model predictive control. PlaNet learns a convolutional variational autoencoder (VAE)~\citep{kingma2013auto} with a pixel reconstruction loss to encode observation into stochastic state representations. The RSSM learns to predict the next stochastic states and environment rewards given previous stochastic and deterministic recurrent states. Following the success of PlaNet on DeepMind Visual Control tasks~\citep{tassa2018deepmind}, Dreamer~\citep{hafner2019dream} improved the algorithm by learning an actor and a value network from the world model representations. DreamerV2~\citep{hafner2020mastering} applied the algorithm to Atari games, utilizing categorical latent states with straight-through gradients~\citep{bengio2013estimating} in the world model to improve performance, instead of Gaussian latents with reparameterized gradients~\citep{kingma2013auto}. DreamerV3~\citep{hafner2023mastering} mastered diverse domains using the same hyper-parameters with a set of architectural changes to stabilize learning across tasks. The agent uses symlog predictions for the reward and value function to address the scale variance across domains. The networks also employ layer normalization~\citep{ba2016layer} to improve robustness and performance while scaling to larger model sizes. It stabilizes policy learning by normalizing the returns and value function using an Exponential Moving Average (EMA) of the returns percentiles. With these modifications, DreamerV3 outperformed specialized model-free and model-based algorithms in a wide range of benchmarks.

In parallel to the Dreamer line of work, \citet{schrittwieser2020mastering} proposed MuZero, a model-based algorithm combining Monte-Carlo Tree Search (MCTS)~\citep{coulom2006efficient} with a powerful world model to achieve superhuman performance in precision planning tasks such as Chess, Shogi and Go. The model is learned by being unrolled recurrently for K steps and predicting environment quantities relevant to planning. The MCTS algorithm uses the learned model to simulate environment trajectories and output an action visit distribution over the root node. This potentially better policy compared to the neural network one is used to train the policy network. More recently, \citet{ye2021mastering} proposed EfficientZero, a sample efficient version of the MuZero algorithm using self-supervised learning to learn a temporally consistent environment model and achieve strong performance on Atari games.

\subsection{Transformer-based World Models}

\begin{table}[!b]
    \caption{Comparison between TWISTER and other recent model-based approaches learning a world model in latent space. \textit{Tokens} refers to tokens used by the autoregressive world model. \textit{Latent} ($z_t$) is image representation while \textit{hidden} ($h_{t}$) is world model hidden state carrying historical information.}
    \centering
    \setlength{\tabcolsep}{2pt}
    \scriptsize
    \hfill \break
    \begin{tabular}{ccccccc}
    \toprule
    Attributes & TWM & IRIS & DreamerV3 & STORM & $\Delta$-IRIS & TWISTER (ours) \\
    \midrule
    World Model & Transformer & Transformer & GRU & Transformer & Transformer & Transformer \\
    Prediction Horizon & Next state & Next state & Next state & Next state & Next state & K = 10 steps \\
    Tokens & Latent, action, reward & Latent ($4 \times 4$) & Latent & Latent & Latent ($2 \times 2$) & Latent \\
    Latent Representation & Categorical-VAE & VQ-VAE & Categorical-VAE & Categorical-VAE & VQ-VAE & Categorical-VAE \\
    Decoder Inputs & Latent & Latent & Latent, hidden & Latent & Latent, action, image & Latent\\
    Agent State ($s_{t}$) & Latent & Image & Latent, hidden & Latent, hidden & Image & Latent, hidden \\
    \bottomrule
    \end{tabular}
    \label{table:world_models}
\end{table}

Recent works have proposed replacing RNN-based world models by Transformer-based architectures using self-attention to process past context. TransDreamer~\citep{chen2022transdreamer} replaced DreamerV3's RSSM by a Transformer State-Space Model (TSSM) using masked self-attention to imagine future trajectories. The agent was evaluated on Hidden Order Discovery tasks requiring long-term memory and reasoning. They also experimented on a few Visual DeepMind Control~\citep{tassa2018deepmind} and Atari~\citep{bellemare2013arcade} tasks, showing comparable performance to DreamerV2. TWM~\citep{robine2023transformer} (Transformer-based World Model) proposed a similar approach, encoding states, actions and rewards as distinct successive input tokens for the autoregressive Transformer. The decoder also reconstructed input images without the world model hidden states, discarding past context temporal information for image reconstruction. More recently, STORM~\citep{zhang2024storm} (Stochastic Transformer-based wORld Model) achieved results comparable to DreamerV3 with better training efficiency on the Atari 100k benchmark. STORM proposed to fuse state and action into a single token for the transformer network compared to TWM which uses distinct tokens. This led to better training efficiency with state-of-the-art performance.

Another line of work focused on designing Transformer-based world model to train agents from reconstructed trajectories in pixel space. Analogously to SimPLe, the agent's policy and value functions are trained from image reconstruction instead of world model hidden state representations. This requires learning auxiliary encoder networks for the policy and value functions. 
Contrary to Dreamer-inspired works that learn agents from world model representations, these approaches also require accurate image reconstruction to train agents effectively. IRIS~\citep{micheli2022transformers} first proposed a world model composed of a VQ-VAE~\citep{van2017neural} to convert input images into discrete tokens and an autoregressive transformer to predict future tokens. IRIS was evaluated on the Atari 100k benchmark~\citep{kaiser2019model} showing promising performance in a low data regime. More recently, \citet{micheli2024efficient} proposed $\Delta$-IRIS, encoding stochastic deltas between time steps using previous action and image as conditions for the encoder and decoder. This increased VQ-VAE compression ratio and image reconstruction capabilities, achieving state-of-the-art performance on the Crafter~\citep{hafner2021benchmarking} benchmark and better results on Atari 100k.

Table~\ref{table:world_models} compares the architectural details of recent model-based approaches learning a world model in latent space with our proposed method. Following preceding Transformer-based approaches, we reconstruct image observation from the encoder stochastic state $z_{t}$ instead of $s_{t}$, which prevents the world model from using temporal information to facilitate reconstruction. The Transformer network uses relative positional encodings~\citep{dai2019transformer}, which simplifies the use of the world model during imagination and evaluation. Absolute positional encodings require the Transformer network to reprocess past latent states with adjusted positional encodings when the current position gets larger than the ones seen during training. We also use the agent state $s_{t}$ as input for predictor networks during the world model training phase to make actor-critic learning more straightforward. 

\subsection{Contrastive Predictive Coding}

Contrastive Predictive Coding (CPC) was introduced by~\citet{oord2018representation} as a representation learning method based on contrastive learning for autoregressive models. CPC encodes a temporal signal into hidden representations and trains an autoregressive model to maximize the mutual information between the autoregressive model output features and future encoded representations using an InfoNCE loss based on Noise-Contrastive Estimation~\citep{gutmann2010noise}. CPC was able to learn useful representations achieving strong performance on four distinct domains: speech phoneme classification, image classification, text classification tasks, and reinforcement learning with DeepMind Lab 3D environments~\citep{beattie2016deepmind}. While CPC was applied to speech, image, and text domains as a pretraining pretext task, it showed promising results on DeepMind Lab tasks being used as an auxiliary loss for the A3C~\citep{mnih2016asynchronous} agent. In this work, we propose to apply CPC to model-based reinforcement learning. We introduce action-conditioned CPC (AC-CPC) that conditions CPC predictions on the sequence of future actions to help the world model to make more accurate predictions and learn higher quality representations. We describe our use of action-conditioned CPC in more detail in section~\ref{subsection:world_model_learning}.

\section{Method}
\label{section:method}

\begin{figure}[ht]
    \centering
    \begin{subfigure}{.59\textwidth}
        \includegraphics[width=\linewidth]{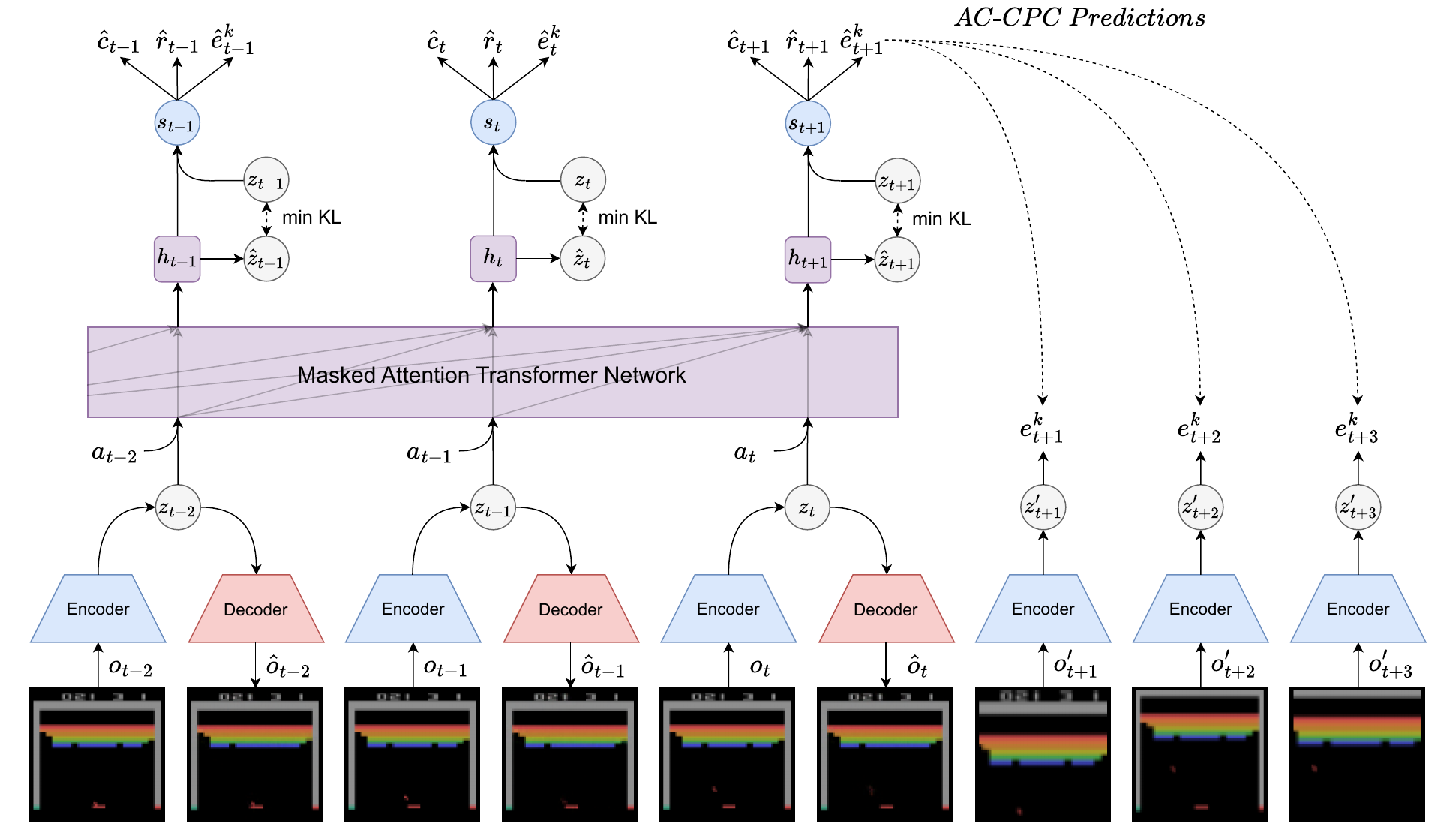}
        \caption{World Model Learning}
        \label{figure:model_1}
    \end{subfigure}\hfill%
    \begin{subfigure}{.40\textwidth}
        \includegraphics[width=\linewidth]{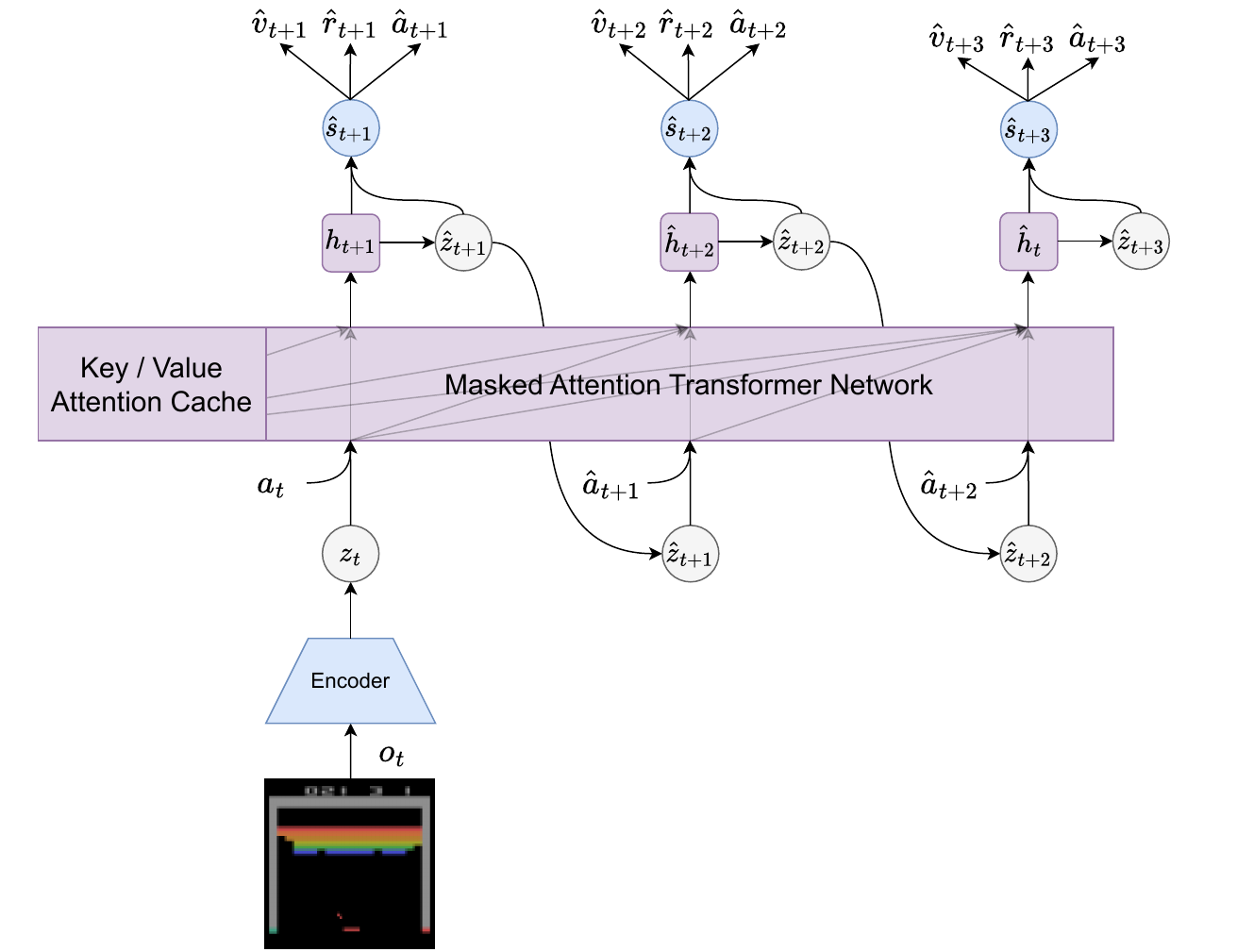}
        \caption{Agent Behavior Learning}
        \label{figure:model_2}
    \end{subfigure}
    \caption{Transformer-based world model with contrastive representations. The world model learns temporal feature representations by maximizing the mutual information between model states $s_{t}$ and future stochastic states $z'_{t:t+K}$ obtained from augmented views of image observations. The encoder network converts image observations into stochastic states $z_{t}$, from which a decoder network learns to reconstruct images while the masked attention Transformer network predicts next episode continuations, rewards and stochastic states conditioned on selected actions.}
    \label{figure:model}
\end{figure}

We introduce TWISTER, a Transformer model-based reinforcement learning algorithm using action-conditioned Contrastive Predictive Coding to learn high-level feature representations and improve the agent performance. TWISTER comprises three main neural networks: a world model, an actor network and a critic network. The world model learns to transform image observations into discrete stochastic states and simulate the environment to generate imaginary trajectories. The actor and critic networks are trained in latent space with imaginary trajectories generated from the world model to select actions maximizing the expected sum of future rewards. The three networks are trained concurrently using a replay buffer sampling sequences of past experiences collected during training. This section describes the architecture and optimization process of our proposed Transformer-based world model with contrastive representations. Analogously to previous approaches, we also detail the learning process of the critic and actor networks taking place in latent space. Figure~\ref{figure:model} shows an overview of our Transformer-based world model trained with AC-CPC. It also illustrates the imagination process undertaken during the agent behavior learning phase.

\subsection{World Model Learning}
\label{subsection:world_model_learning}

Consistent with prior works~\citep{hafner2023mastering, robine2023transformer, zhang2024storm}, we learn a world model in latent space by encoding input image observations $o_{t}$ into hidden representations using a convolutional VAE with categorical latents. The hidden representations are linearly projected to categorical distribution logits comprising 32 categories, each with 32 classes, from which discrete stochastic states $z_{t}$ are sampled. The world model is implemented as a Transformer State-Space Model (TSSM)~\citep{chen2022transdreamer} using masked self-attention to predict next stochastic states $\hat{z}_{t+1}$ given previous states $z_{1:t}$ and actions $a_{1:t}$. The Transformer network outputs hidden states $h_{t}$ that are concatenated with stochastic states $z_{t}$ to form the model states $s_{t}=\{h_{t}, z_{t}\}$. The world model predicts environment reward $\hat{r}_{t}$, episode continuation $\hat{c}_{t}$ and AC-CPC features $\hat{e}^{k}_{t}$ using simple Multi Layer Perceptron (MLP) networks. The trainable world model components are the following:
\begin{equation}
\setlength{\tabcolsep}{1pt}
\renewcommand{\arraystretch}{1.2}
\begin{tabular}{rllrl}
    \raisebox{1.6ex}{\multirow{4}{*}{TSSM $\begin{dcases} \\ \\ \\ \end{dcases}$}} & Encoder Network: & \hspace{1em} & $z_{t}$ & $\sim q_{\phi}(z_{t}\ |\ o_{t})$  \\
    & Transformer Network: & & $h_{t}$ & $= f_{\phi}(z_{1:t-1}, a_{1:t-1})$    \\
    & Dynamics Predictor: & & $\hat{z}_{t}$ & $\sim p_{\phi}(\hat{z}_{t}\ |\ h_{t})$  \\
    & Decoder Network: & & $\hat{o}_{t}$ & $\sim p_{\phi}(\hat{o}_{t}\ |\ z_{t})$\\
    & Reward Predictor: & & $\hat{r}_{t}$ & $\sim p_{\phi}(\hat{r}_{t}\ |\ s_{t})$ \\
    & Continue Predictor: & & $\hat{c}_{t}$ & $\sim p_{\phi}(\hat{c}_{t}\ |\ s_{t})$ \\
    \raisebox{1.6ex}{\multirow{3}{*}{AC-CPC $\begin{dcases} \\ \\ \end{dcases}$}} & Representation Network: & & $e^{k}_{t}$ & $= q^{k}_{\phi}(z'_{t+k})$\\
    & AC-CPC Predictor: & & $\hat{e}^{k}_{t}$ & $= p^{k}_{\phi}(s_{t}, a_{t:t+k})$\\
\end{tabular}
\end{equation}

\paragraph{Transformer State-Space Model}

We train an autoregressive Transformer network using masked self-attention with relative positional encodings~\citep{dai2019transformer}. During both training, exploration and evaluation, the hidden state sequence computed for the previous segment or state is cached to be reused as an extended context when the model processes the next state. This encoding and caching mechanism allows the world model to imagine future trajectories from any state, eliminating the need to reprocess latent states with adjusted positional encodings.

\paragraph{World model losses} 

Given an input batch containing $B$ sequences of $T$ image observations $o_{1:T}$, actions $a_{1:T}$ , rewards $r_{1:T}$ , and episode continuation flags $c_{1:T}$, the world model parameters ($\phi$) are optimized to minimize the following loss function:
\begin{equation}
L(\phi) = \frac{1}{BT}\sum_{b=1}^{B}\sum_{t=1}^{T}\Bigr[L_{rew}(\phi) + L_{con}(\phi) + L_{rec}(\phi) + L_{dyn}(\phi) + L_{cpc}(\phi)\Bigl]
\end{equation}
$L_{rew}$ and $L_{con}$ train the world model to predict environment rewards and episode continuation flags, which are used to compute the returns of imagined trajectories during the behavior learning phase. We adopt the symlog cross-entropy loss from DreamerV3~\citep{hafner2023mastering}, which scales and transforms rewards into twohot encoded targets to ensure robust learning across games with different reward magnitudes. The reconstruction loss $L_{rec}$ trains the categorical VAE to learn stochastic representations $z_{t}$ for the world model by reconstructing input visual observations $o_{t}$:
\begin{subequations}
\begin{align}
L_{rew}(\phi) &= \operatorname{SymlogCrossEnt}(\hat{r}_{t}, r_{t}) \\
L_{con}(\phi) &= \operatorname{BinaryCrossEnt}(\hat{c}_{t}, c_{t}) \\
L_{rec}(\phi) &= ||\hat{o}_{t} - o_{t}||_{2}^{2}
\end{align}
\end{subequations}
The world model dynamics loss $L_{dyn}$ trains the dynamics predictor network to predict the next stochastic states representations from transformer hidden states by minimizing the Kullback–Leibler (KL) divergence between the predictor output distribution $p_{\phi}(\hat{z}_{t}\ |\ h_{t})$ and the next encoder representation $q_{\phi}(z_{t}\ |\ o_{t})$. We also add a regularization term to avoid spikes in the KL loss and stabilize learning by training the encoder representations to become more predictable. Both loss terms use the stop gradient operator $sg(\cdot)$ to prevent the gradients of targets from being backpropagated and are scaled with loss weights $\beta_{dyn}=0.5$ and $\beta_{reg}=0.1$, respectively:
\begin{equation}
\setlength{\tabcolsep}{2pt}
\renewcommand{\arraystretch}{1.5}
\begin{tabular}{rll}
$L_{dyn}(\phi)=$ & $\beta_{dyn}$ & $\max\bigl(1, \text{KL}\bigl[\ sg( q_{\phi}(z_{t}\ |\ o_{t}))\ ||\hspace{3.6ex}p_{\phi}(\hat{z}_{t}\ |\ h_{t})\hspace{0.87ex}\bigr]\bigr)$ \\
 $+$ & $\beta_{reg}$ & $\max\bigl(1, \text{KL}\bigl[\hspace{3.6ex}q_{\phi}(z_{t}\ |\ o_{t})\hspace{1.42ex} ||\  sg(  p_{\phi}(\hat{z}_{t}\ |\ h_{t}) )\bigr]\bigr)$
\end{tabular}
\end{equation}
The Transformer network learns feature representations using action-conditioned Contrastive Predictive Coding. The representations are learned by maximizing the mutual information between model states $s_{t}$ and future stochastic states $z'_{t:t+K}$ obtained from augmented views of image observations. We adopt a simple strategy to generate negative samples: Given the sequence batch of augmented stochastic states $Z'$ containing one positive sample, we treat the other $B \times T -1$ samples as negatives. The world model learns to distinguish positive samples from negatives using InfoNCE:
\begin{equation}
L_{cpc}(\phi) = - \frac{1}{K} \sum_{k=0}^{K-1} \log \frac{exp(sim(z'_{t+k}, s_{t}))}{\sum_{z'_{j} \in Z'} exp(sim(z'_{j}, s_{t}))}
\label{equation:loss_cpc}
\end{equation}
The world model learns to predict $K=10$ future stochastic states among the batch of augmented samples. We compute similarities as dot products: $sim(z'_{j}, s_{t})=q_{\phi}^{k}(z'_{j})^{T}p_{\phi}^{k}(s_{t}, a_{t:t+k})$, learning two MLP networks  $q_{\phi}^{k}$ and $p_{\phi}^{k}$ for each step k. Contrary to the original CPC paper, which experiments with continuous feature states, we use discrete latent states for the world model. This requires learning a representation network $q_{\phi}^{k}$ to project discretized stochastic states $z'_{j}$ to contrastive feature representations $e_{t}^{k}$. The AC-CPC predictor $p_{\phi}^{k}$ uses the concatenated sequence of future actions $a_{t:t+k}$ as condition to reduce uncertainty and learn quality representations.

\begin{figure}[t!]
        \centering
        \includegraphics[width=0.85\linewidth]{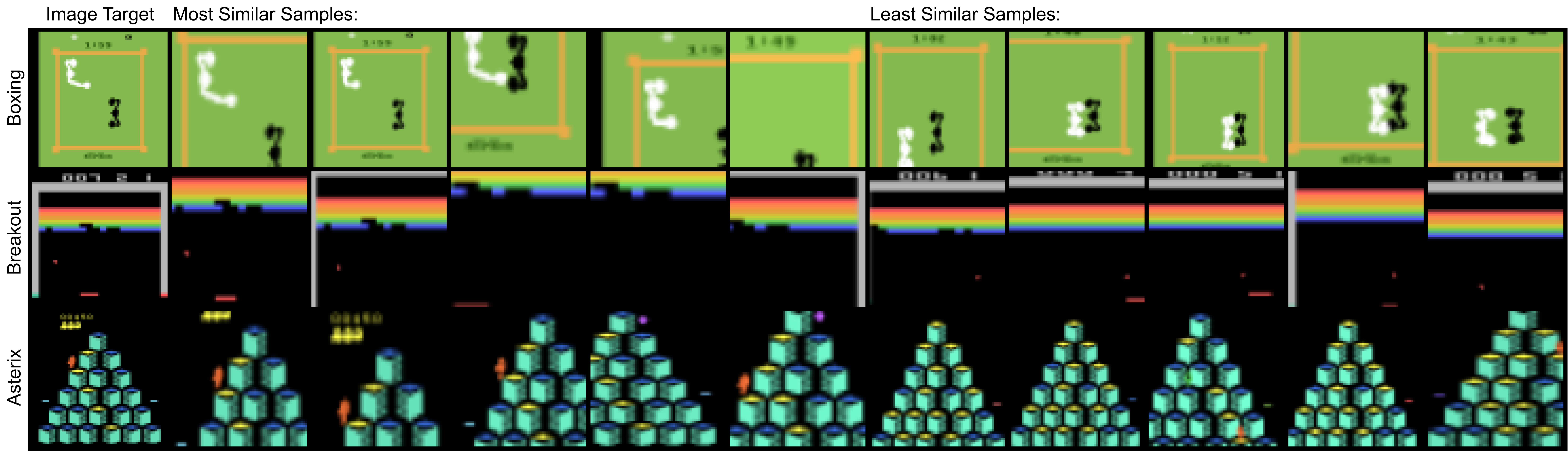}
        \caption{AC-CPC predictions made by the world model. We show the target positive sample without augmentation and the predicted most/least similar samples among the batch of augmented image views. We observe that TWISTER learns to identify most/least similar samples to the future target state using observation details such as the ball position, game score or agent movements. AC-CPC necessitates the agent to focus on observation details to accurately predict future samples, thereby preventing common failure cases where small objects are ignored by the reconstruction loss.}
        \label{figure:cpc_preds_small}
        \vspace{-2ex}
\end{figure}

\subsection{Agent Behavior Learning}

The agent critic and actor networks are trained with imaginary trajectories generated from the world model. In order to compare TWISTER with previous approaches that train agents using world model representations, we adopt the agent behavior learning settings from DreamerV3~\citep{hafner2023mastering}. Learning takes place entirely in latent space, which allows the agent to process large batch sizes and increase generalization. We flatten the model states of the sampled sequences along the batch and time dimensions to generate $B^{img}=B \times T$ sample trajectories using the world model. The self-attention keys and values features computed during the world model training phase are cached to be reused during the agent behavior learning phase and preserve past context. As shown in Figure~\ref{figure:model_2}, the world model imagines $H=15$ steps into the future using the Transformer network and the dynamics network head, selecting actions by sampling from the actor network categorical distribution. Analogously to world model predictor networks, the actor and critic networks are designed as simple MLPs with parameter vectors ($\theta$) and ($\psi$), respectively. 
\begin{equation}
\setlength{\tabcolsep}{10pt}
\begin{tabular}{ll}
    Actor Network: & $a_{t} \sim \pi_{\theta}(a_{t} | s_{t})$ \\
    Critic Network: & $v_{t} \sim V_{\psi}(v_{t} | s_{t})$ 
\end{tabular}
\end{equation}

\paragraph{Critic Learning}

Following DreamerV3, the critic network learns to minimize the symlog cross-entropy loss with discretized $\lambda$-returns obtained from imagined trajectories with rewards and episode continuation flags predicted by the world model: 
\begin{equation}
\setlength{\tabcolsep}{10pt}
\begin{tabular}{ll}
$R_{t}^{\lambda} = \hat{r}_{t+1} + \gamma \hat{c}_{t+1} \Bigl((1-\lambda)V_{\psi}(s_{t+1}) + \lambda R_{t+1}^{\lambda}\Bigr)$ & $R_{H+1}^{\lambda} = V_{\psi}(s_{H+1})$ 
\end{tabular}
\end{equation}
The critic does not use a target network but relies on its own predictions for estimating rewards beyond the prediction horizon. This requires stabilizing the critic by adding a regularizing term toward the outputs of its own EMA network $V_{\psi'}$. Equation~\ref{equation:critic_loss} defines the critic network loss:
\begin{equation}
L_{critic}(\psi) = \frac{1}{BH}\sum_{b=1}^{B}\sum_{t=1}^{H}\Bigl[\ \underbracket[0.1ex]{\operatorname{SymlogCrossEnt}\bigl(v_{t}, R_{t}^{\lambda}\bigr)}_{\text{discrete returns regression}} + \underbracket[0.1ex]{\operatorname{SymlogCrossEnt}\bigl(v_{t}, V_{\psi'}(s_{t})\bigr)}_{\text{critic EMA regularizer}}\ \Bigr]
\label{equation:critic_loss}
\end{equation}

\paragraph{Actor Learning}

The actor network learns to select actions that maximize the predicted returns using Reinforce~\citep{williams1992simple} while maximizing the policy entropy to ensure sufficient exploration during both data collection and imagination. The actor network loss is defined as follows:
\begin{equation}
L_{actor}(\theta) = \frac{1}{BH}\sum_{b=1}^{B}\sum_{t=1}^{H}\Bigl[\ \underbracket[0.1ex]{-\ sg(A_{t}^{\lambda}) \log \pi_{\theta}(a_{t}\ |\ s_{t})}_{\text{reinforce}} \underbracket[0.1ex]{-\ \eta \mathrm{H}\bigl(\pi_{\theta}(a_{t}\ |\ s_{t})\bigr)}_{\text{entropy regularizer}}\ \Bigr]
\end{equation}
Where $A_{t}^{\lambda}=\big(\hat{R}_{t}^{\lambda} - V_{\psi}(s_{t})\big) / \max(1, S)$ defines advantages computed using normalized returns. The returns are scaled using exponentially moving average statistics of their $5^{th}$ and $95^{th}$ batch percentiles to ensure stable learning across all Atari games:
\begin{equation}
S=\operatorname{EMA}(\operatorname{Per}(R_{t}^{\lambda}, 95) - \operatorname{Per}(R_{t}^{\lambda}, 5), momentum=0.99)
\end{equation}
\section{Experiments}

In this section, we describe our experiments on the commonly used Atari 100k benchmark. We compare TWISTER with SimPLe, DreamerV3 and recent Transformer model-based approaches in Table~\ref{table:results_atari100k}. We also perform several ablation studies on the principal components of TWISTER.

\subsection{Atari 100k Benchmark}

The Atari 100k benchmark was proposed in~\citet{kaiser2019model} to evaluate reinforcement learning agents on Atari games in low data regime. The benchmark includes 26 Atari games with a budget of 400k environment frames, amounting to 100k interactions between the agent and the environment using the default action repeat setting. This amount of environment steps corresponds to about two hours (1.85 hours) of real-time play, representing a similar amount of time that a human player would need to achieve reasonably good performance. The current state-of-the-art is held by EfficientZero V2~\citep{wang2024efficientzero}, which uses Monte-Carlo Tree Search to select the best action at every time step. Another recent notable work is BBF~\citep{schwarzer2023bigger}, a model-free agent using learning techniques that are orthogonal to our work such as periodic network resets and hyper-parameters annealing to improve performance. In this work, to ensure fair comparison and demonstrate the effectiveness of AC-CPC for learning world models, we compare our method with model-based approaches that do not utilize look-ahead search techniques. Combining these additional components with TWISTER would nevertheless be an interesting research direction for future works.

\subsection{Results}

\begin{table}[ht]
    \caption{Agent scores and human-normalized metrics on the 26 games of the Atari 100k benchmark. We show average scores over 5 seeds. Bold numbers indicate best performing method for each game.}
    \centering
    \setlength{\tabcolsep}{5pt}
    \scriptsize
    \hfill \break
    \begin{tabular}{lrrrrrrrrr}
    \toprule
    Game & Random & Human & SimPLe & TWM & IRIS & DreamerV3 & STORM & $\Delta$-IRIS & TWISTER (ours) \\ 
    \midrule
    Alien & 228 & 7128  & 617 & 675 & 420 & 959 & \textbf{984} & 391 & 970 \\
    Amidar & 6 & 1720  & 74 & 122 & 143 & 139 & \textbf{205} & 64 & 184 \\
    Assault & 222 & 742  & 527 & 683 & \textbf{1524} & 706 & 801 & 1123 & 721 \\
    Asterix & 210 & 8503  & 1128 & 1116 & 854 & 932 & 1028 & \textbf{2492} & 1306 \\
    Bank Heist & 14 & 753 & 34 & 467 & 53 & 649 & 641 & \textbf{1148} & 942 \\
    Battle Zone & 2360 & 37188 & 4031 & 5068 & 13074 & 12250 & \textbf{13540} & 11825 & 9920 \\
    Boxing & 0 & 12 & 8 & 78 & 70 & 78 & 80 & 70 & \textbf{88} \\
    Breakout & 2 & 30 & 16 & 20 & 84 & 31 & 16 & \textbf{302} & 35\\
    Chopper Command & 811 & 7388 & 979 & 1697 & 1565 & 420 & \textbf{1888} & 1183 & 910 \\
    Crazy Climber & 10780 & 35829 & 62584 & 71820 & 59324 & \textbf{97190} &  66776 & 57854 & 81880 \\
    Demon Attack & 152 & 1971 & 208 & 350 & \textbf{2034} & 303 & 165 & 533 & 289 \\
    Freeway & 0 & 30 & 17 & 24 & 31 & 0 &  \textbf{34} & 31 & 32 \\
    Frostbite & 65 & 4335 & 237 & \textbf{1476} & 259 & 909 & 1316 & 279 & 305  \\
    Gopher & 258 & 2412 & 597 & 1675 & 2236 & 3730 & 8240 & 6445 & \textbf{22234} \\
    Hero & 1027 & 30826 & 2657 & 7254 & 7037 & \textbf{11161} & 11044 & 7049 & 8773 \\
    James Bond & 29 & 303 & 100 & 362 & 463 & 445 & 509 & 309 & \textbf{573} \\
    Kangaroo & 52 & 3035 & 51 & 1240 & 838 & 4098 & 4208 & 2269 & \textbf{6016} \\
    Krull & 1598 & 2666 & 2205 & 6349 & 6616 & 7782 & 8413 & 5978 & \textbf{8839} \\
    Kung Fu Master & 258 & 22736 & 14862 & 24555 & 21760 & 21420 & \textbf{26182} & 21534 & 23442 \\
    Ms Pacman & 307 & 6952 & 1480 & 1588 & 999 & 1327 & \textbf{2673}  & 1067 & 2206 \\
    Pong & –21 & 15 & 13 & 19 & 15 & 18 & 11 & \textbf{20} & \textbf{20} \\
    Private Eye & 25 & 69571 & 35 & 87 & 100 & 882 & \textbf{7781} & 103 & 1608\\ 
    Qbert & 164 & 13455 & 1289 & 3331 & 746 & 3405 & \textbf{4522} & 1444 & 3197\\
    Road Runner & 12 & 7845 & 5641 & 9107 & 9615 & 15565 & 17564 & 10414 & \textbf{17832} \\ 
    Seaquest & 68 & 42055 & 683 & \textbf{774} & 661 & 618 & 525 & 827 & 532 \\
    Up N Down & 533 & 11693 & 3350 & \textbf{15982} & 3546 & 7600 & 7985 & 4072 & 7068\\
    \midrule
    \# Superhuman & 0 & N/A & 1 & 8 & 10 & 9 & 10 & 11 & \textbf{12} \\
    Normed Mean (\%) & 0 & 100 & 33  & 96 & 105 & 112 & 127 & 139 & \textbf{\resultmean}\\
    Normed Median (\%) & 0 & 100 & 13 & 51 & 29 & 49 & 58 & 53 & \textbf{\resultmedian} \\
    \bottomrule
    \end{tabular}
    \label{table:results_atari100k}
\end{table}

Table~\ref{table:results_atari100k} compares TWISTER with SimPLe~\citep{kaiser2019model}, DreamerV3~\citep{hafner2023mastering} and recent Transformer model-based approaches~\citep{robine2023transformer, micheli2022transformers, zhang2024storm, micheli2024efficient} on the Atari 100k benchmark. Following preceding works, we use human-normalized metrics and compare the mean and median returns across all 26 games. The human-normalized scores are computed for each game using the scores achieved by a human player and the scores obtained by a random policy: $normed\ score=\frac{agent\ score - random\ score}{human\ score - random\ score}$. We also show stratified bootstrap confidence intervals of the human-normalized mean and median in Figure~\ref{figure:CI}. Performance curves corresponding to individual games can be found in the appendix~\ref{appendix:results_atari100k}.

\begin{wrapfigure}{r}{0.48\textwidth}
    \centering
    \vspace{-2ex}
    \includegraphics[width=\linewidth]{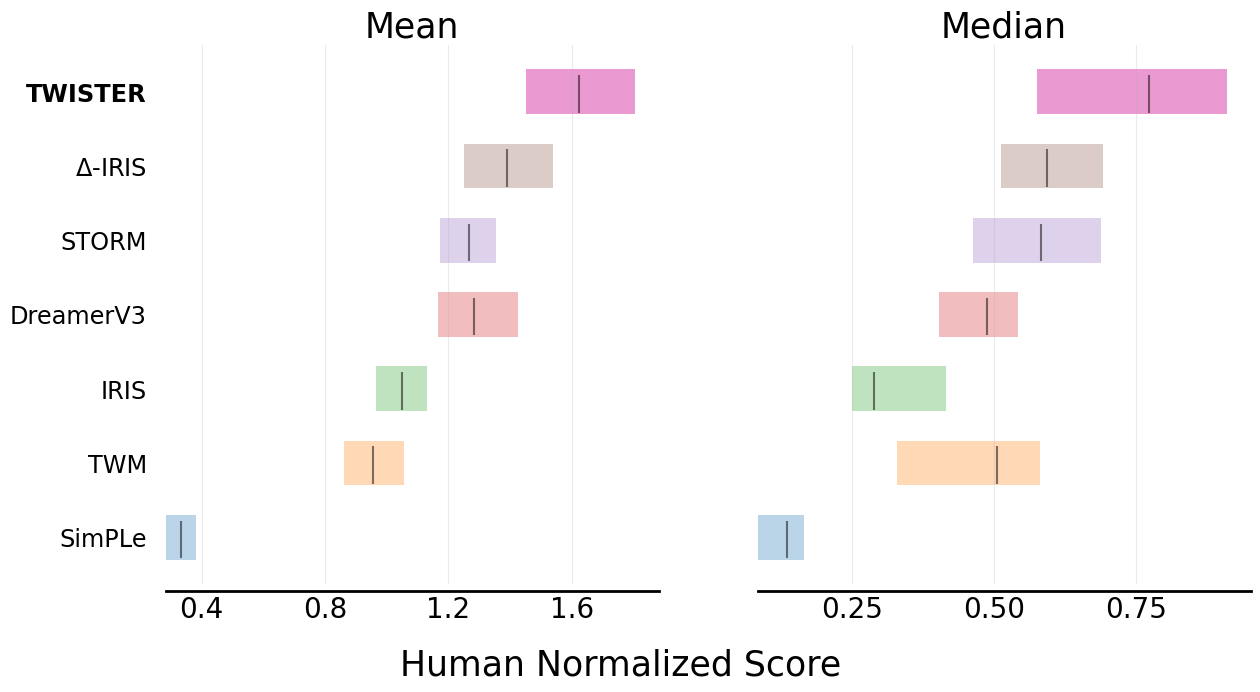}
    \caption{Mean and median scores, computed with stratified bootstrap confidence intervals~\citep{agarwal2021deep}. TWISTER achieves a normalized mean of 1.62 and a median of 0.77.}
    \vspace{-1ex}
    \label{figure:CI}
\end{wrapfigure}

TWISTER achieves a human-normalized mean score of \resultmean\% and a median of \resultmedian\% on the Atari 100k benchmark, setting a new record among state-of-the-art model-based methods that do not employ look-ahead search techniques. Analogously to STORM, we find that TWISTER demonstrates superior performance in games where key objects related to rewards are numerous, such as \textit{Amidar}, \textit{Bank Heist}, \textit{Gopher} and \textit{Ms Pacman}. Furthermore, we observe that TWISTER benefits from increased performance in games with small moving objects like \textit{Breakout}, \textit{Pong} and \textit{Asterix}. We suppose that the AC-CPC objective requires the agent to focus on the ball's position in these games to accurately predict future samples, thereby preventing failure cases where small objects are ignored by the reconstruction loss. Alternatively, IRIS and $\Delta$-IRIS solve this issue by learning agents from high-quality reconstructed images. They encode image observations into spatial latent spaces through a VQ-VAE structure, which allows these approaches to better capture details and achieve lower reconstruction errors with good results for these games. We show CPC predictions made by the world model for diverse Atari games in the appendix~\ref{appendix:cpc_preds}.

\subsection{Ablation Studies}

In order to study the impact of AC-CPC on TWISTER performance, we perform ablation studies on all 26 games of the Atari 100k benchmark, applying one modification at a time. We experiment with the number of CPC steps predicted by the world model. We show that data augmentation helps to complexify the AC-CPC objective and improve its effectiveness. We find that conditioning CPC predictions on the sequence of future actions leads to more accurate predictions and improves the quality of representations. We also study the effect of world model design on AC-CPC effectiveness. Table~\ref{table:ablations} shows the aggregated scores obtained for the main ablations after 400k environment steps.

\begin{table}[!ht]
    \caption{Ablations of the AC-CPC loss, contrastive samples augmentation, conditioning on future actions and using DreamerV3's RSSM. We perform one modification at a time and evaluate on the 26 Atari games. The detailed results obtained for individual games can be found in the appendix~\ref{appendix:main_ablations}.}
    \centering
    \setlength{\tabcolsep}{5pt}
    \scriptsize
    \hfill \break
    \begin{tabular}{lccccc}
    \toprule
    Metrics & TWISTER & No AC-CPC & DreamerV3 World Model & No Action Conditioning & No Data Augmentation \\
    \midrule
    Normed Mean (\%) & \textbf{\resultmean} & 112 & 121 & 111 & 120 \\
    Normed Median (\%) & \textbf{\resultmedian} & 44 & 69 & 42 & 68 \\
    \bottomrule
    \end{tabular}
    \label{table:ablations}
    \vspace{-3ex}
\end{table}

\begin{figure}[b!]
    \vspace{-2ex}
    \centering
    \begin{subfigure}{0.5\textwidth}
        \centering
        \includegraphics[width=0.775\linewidth]{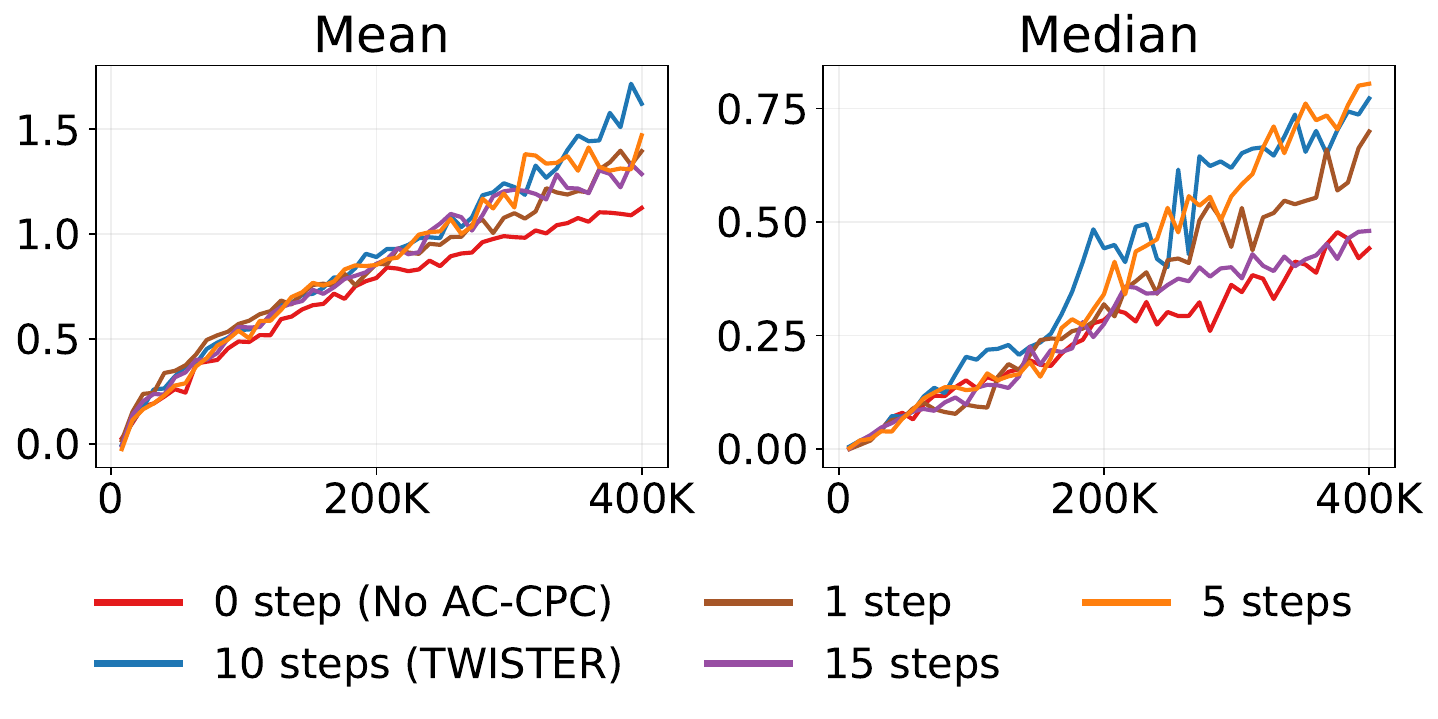}
        \caption{Number of Contrastive steps}
        \label{figure:ablations_cpc_steps}
    \end{subfigure}\hfill%
    \begin{subfigure}{0.5\textwidth}
        \centering
        \includegraphics[width=0.775\linewidth]{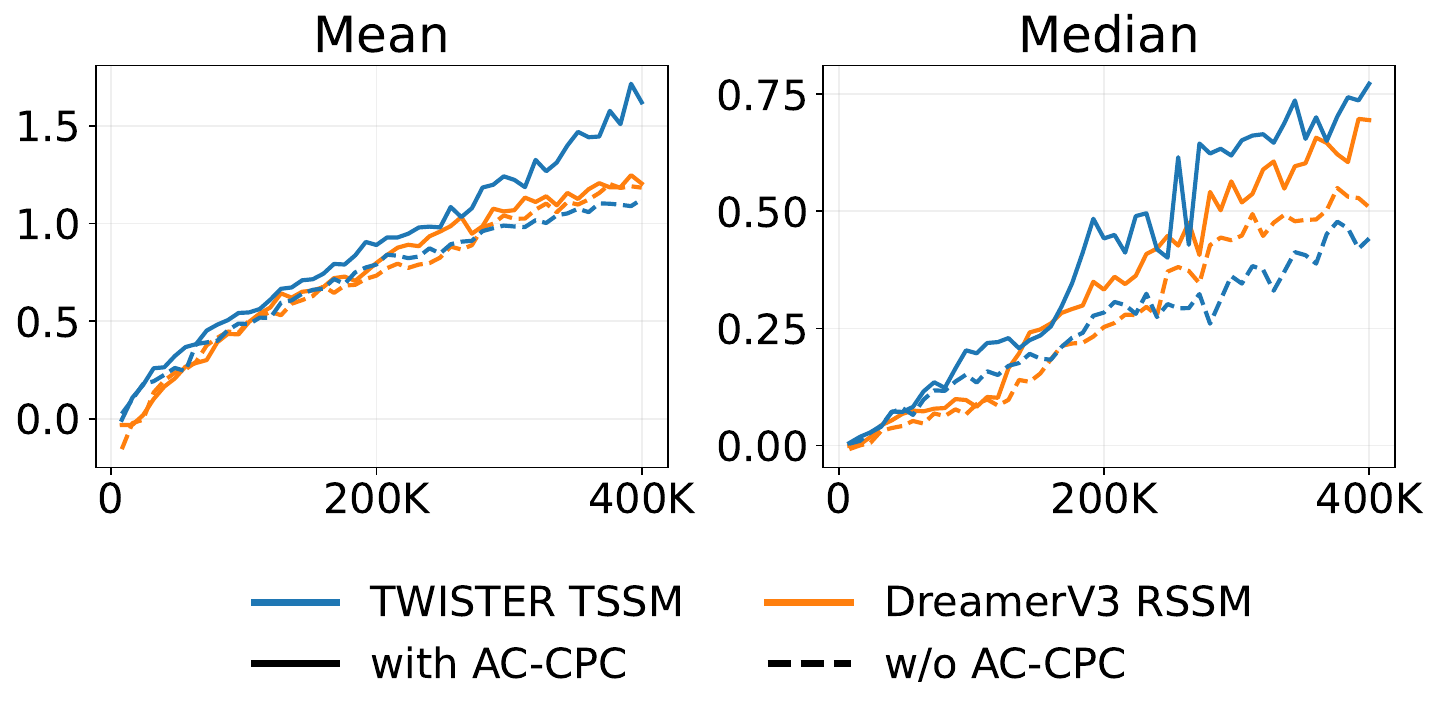}
        \caption{World Model Architecture}
        \label{figure:ablations_dreamer_twister_cpc}
    \end{subfigure} 
    \\ \vspace{1ex}
    \begin{subfigure}{0.5\textwidth}
        \centering
        \includegraphics[width=0.775\linewidth]{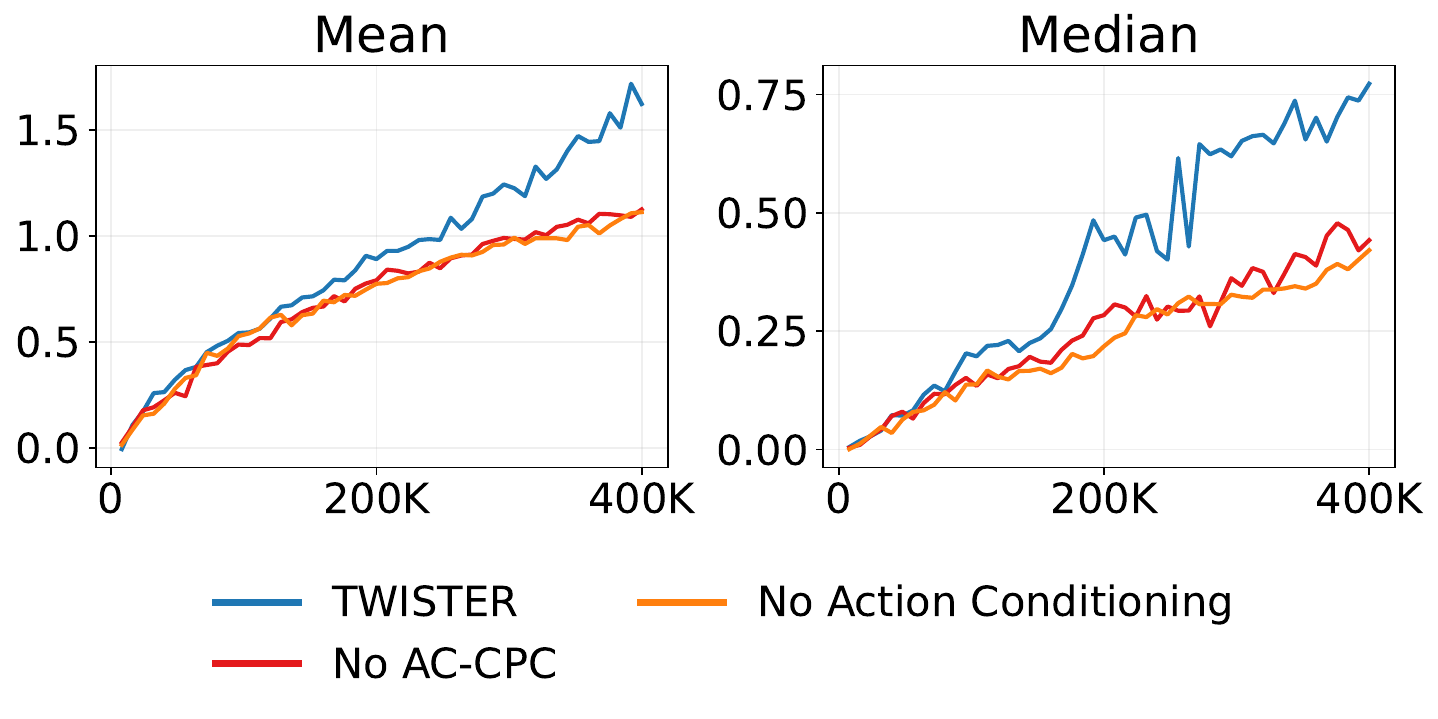}
        \caption{Action Conditioning}
        \label{figure:ablations_action_cond}
    \end{subfigure}\hfill%
    \begin{subfigure}{0.5\textwidth}
        \centering
        \includegraphics[width=0.775\linewidth]{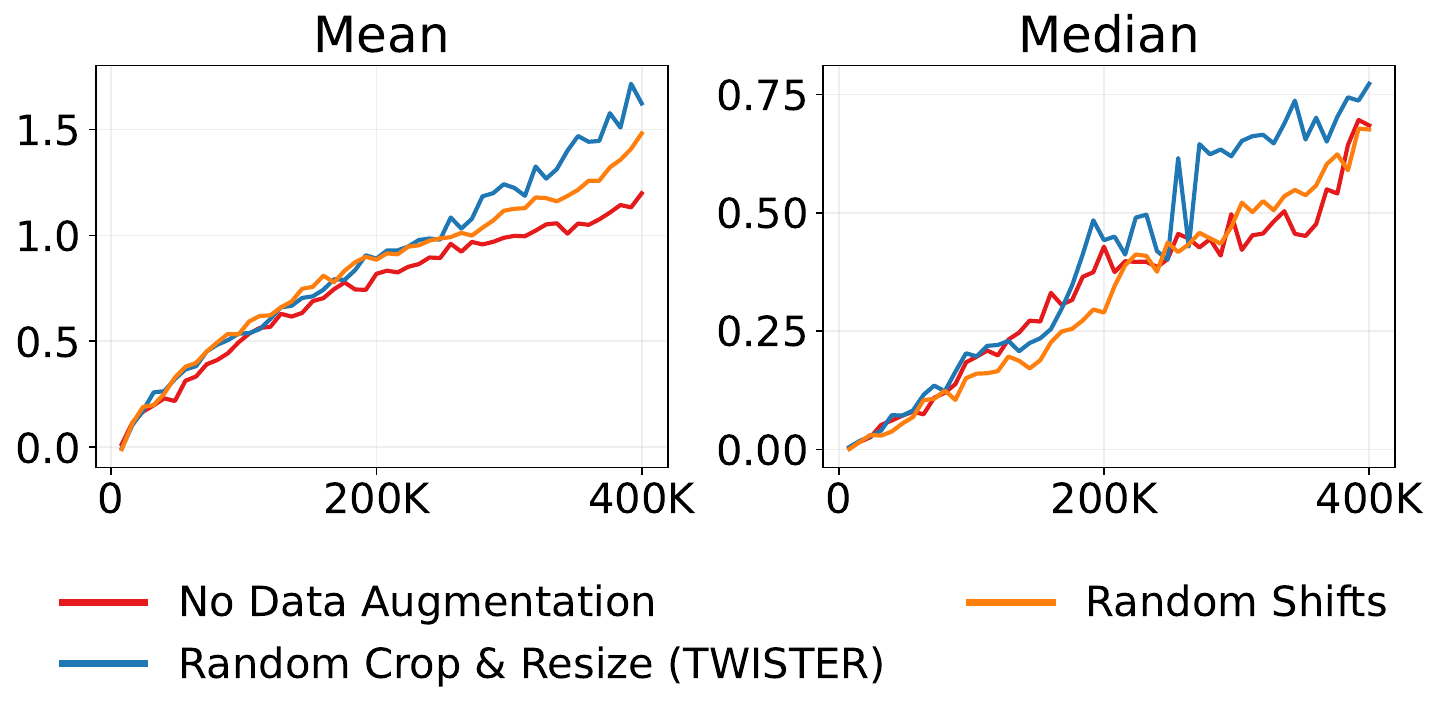}
        \caption{Data Augmentation}
        \label{figure:ablations_augments}
    \end{subfigure}
    \caption{Ablations made on the Atari 100k benchmark. The results are averaged over 5 seeds. We study the effect of data augmentation, action conditioning and the number of predicted CPC steps on TWISTER performance. We also study the effect of world model design on AC-CPC effectiveness.}
    \label{figure:ablations}
\end{figure}

\paragraph{Number of Contrastive Steps} We experiment with several numbers of CPC steps, comparing human-normalized metrics over all 26 games of the Atari100k benchmark. Figure~\ref{figure:ablations_cpc_steps} shows that TWISTER achieves the best human-normalized mean score when predicting 10 steps into the future, corresponding to 0.67 seconds of game time. We find that AC-CPC has a significant effect on TWISTER performance up to a certain amount of steps. We observe an increase in human-normalized mean and median scores with the number of predicted CPC steps. However, a degradation of the results is noticed when predicting 15 steps into the future. The difference in median score indicates a decrease in performance for middle-scoring games.

\paragraph{World Model Architecture} 

We study the impact of world model design on AC-CPC effectiveness to learn feature representations. Figure~\ref{figure:ablations_dreamer_twister_cpc} shows the effect of AC-CPC on performance when replacing the TSSM of TWISTER with DreamerV3's RSSM~\citep{hafner2023mastering}. While the two approaches achieve similar results without the AC-CPC objective, we find that AC-CPC has a significant effect on TWISTER, improving performance on most games. These findings can be attributed to the fact that Transformers are generally more effective than RNNs at learning feature representations due to several key architectural differences. The capacity of self-attention to model temporal relationships without recurrence makes the Transformer architecture highly effective at capturing context and learning hierarchical features. On the other hand, the recurrent nature of RNNs can lead to vanishing gradients and slower convergence, particularly with long sequences.

\begin{wrapfigure}{r}{0.48\textwidth}
    \centering
    \vspace{0.5ex}
    \includegraphics[width=\linewidth]{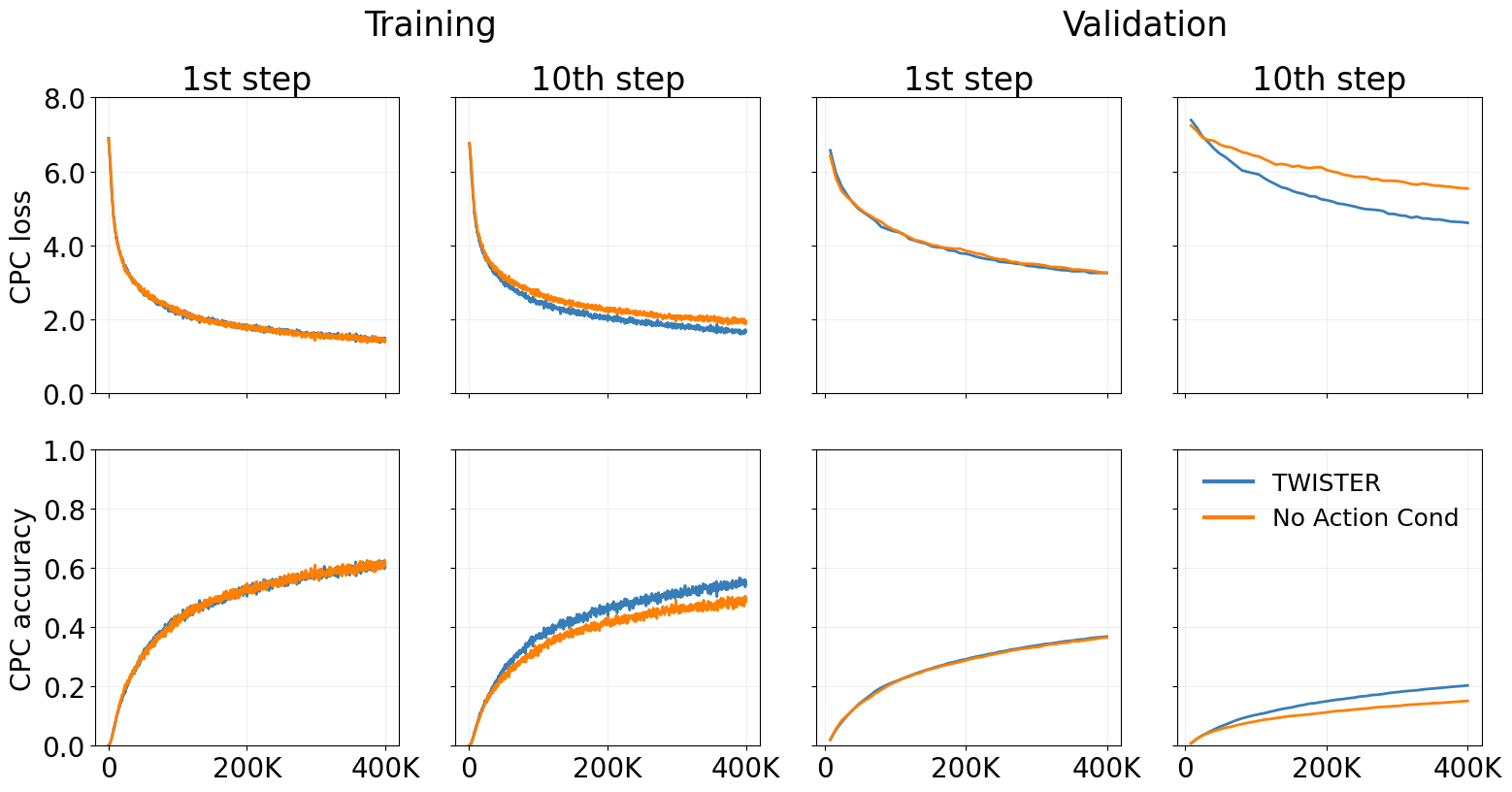}
    \vspace{-3ex}
    \caption{Aggregated CPC loss and prediction accuracy over all 26 games. We use a validation replay buffer of 100k samples to compare CPC loss on unseen trajectories. The trajectories are obtained from a collection of DreamerV3 and TWISTER agents pretrained with 5 seeds.}
    \vspace{-1ex}
    \label{figure:action_cond_cpc_loss_acc}
\end{wrapfigure}

\paragraph{Actions Conditioning} We find that conditioning the CPC prediction head on the sequence of future actions leads to more accurate predictions and higher quality representations. Figure~\ref{figure:action_cond_cpc_loss_acc} shows the aggregated CPC loss and prediction accuracy for training and validation sequences over all Atari games. We report the average number of times the similarity for the positive sample is higher than for the negative samples in the contrastive loss. Without knowing the sequence of future actions, the world model cannot predict future environment states accurately, which makes the task almost insolvable and counterproductive beyond a certain amount of CPC steps. We observe a decrease in accuracy compared to TWISTER when predicting multiple steps without knowing the sequence of future actions. Figure~\ref{figure:ablations_action_cond} shows the aggregated human-normalized scores over the 26 games when removing the condition of future actions for CPC predictions. We find that the CPC objective does not bring notable performance improvements when removing future actions conditioning.

\begin{wrapfigure}{r}{0.48\textwidth}
    \centering
    \vspace{-3ex}
    \includegraphics[width=\linewidth]{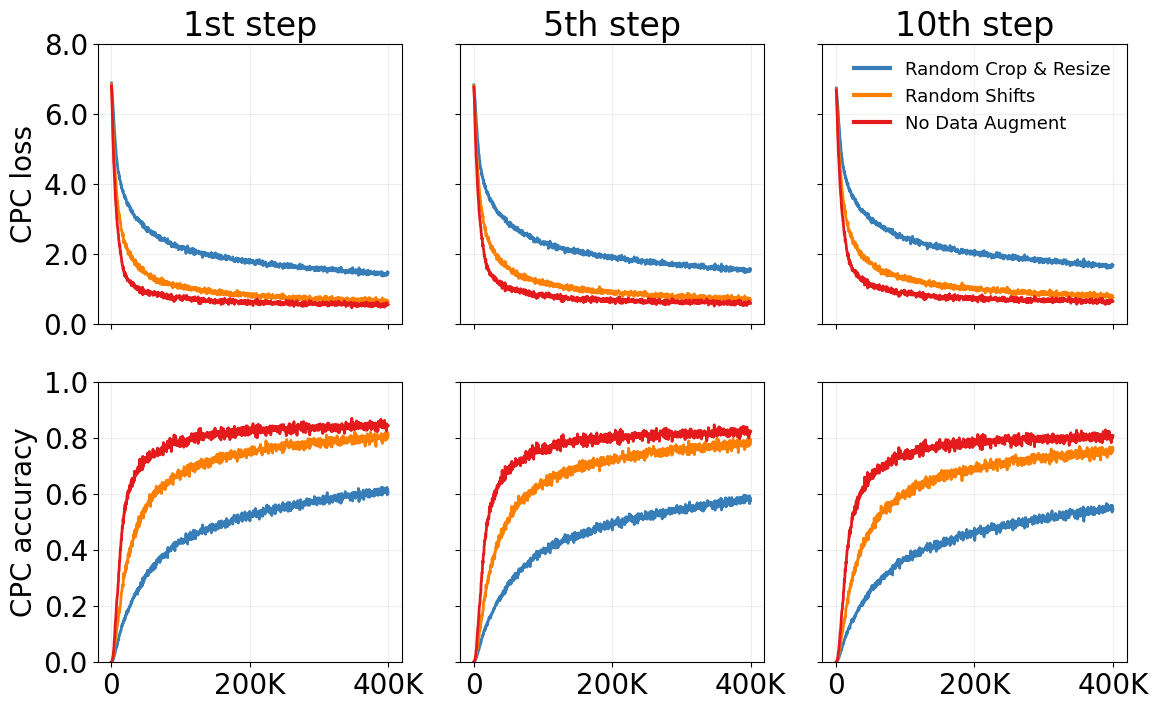}
    \vspace{-3ex}
    \caption{Effect of data augmentation on AC-CPC objective complexity. We aggregate CPC loss and prediction accuracy over all Atari games for different time horizons.}
    \vspace{-3ex}
    \label{figure:augments_cpc_loss_acc}
\end{wrapfigure}

\paragraph{Effect of Data Augmentation} 

The effect of data augmentation on CPC performance was studied by \citet{kharitonov2021data}. In their work, they propose to introduce data augmentation for CPC to learn higher quality speech representations, yielding better performances. In this work, we apply image augmentation to contrastive samples in order to complexify the AC-CPC objective and make the representation learning task more challenging. We apply the commonly used \textit{random crop and resize} augmentation during training for its effectiveness in the area of image-based contrastive learning~\citep{chen2020simple}. The use of random crops requires the world model to identify several key elements in the observations in order to accurately predict positives samples. We also experiment with \textit{random shifts}~\citep{yarats2021image}, shifting the image up to 4 pixels in height and width but found it to have a lesser impact on the learning objective. Figure~\ref{figure:ablations_augments} shows the aggregated human-normalized scores for studied augmentation techniques. We find that \textit{random crop and resize} helps the best to improve final performance. Not using image augmentations for negative and positive samples reduces the impact of AC-CPC on TWISTER performance, achieving lower mean and median scores. We show the impact of data augmentation on the AC-CPC objective complexity for different time horizons in Figure~\ref{figure:augments_cpc_loss_acc}. 

\section{Conclusion}

We propose TWISTER, a Transformer model-based reinforcement learning agent learning high-level temporal feature representations with action-conditioned Contrastive Predictive Coding. TWISTER achieves new state-of-the-art results on the Atari 100k benchmark among model-based approaches that do not employ look-ahead search with a human-normalized mean and median score of \resultmean\% and \resultmedian\%, respectively. We study the impact of learning contrastive representations on Transformer-based world models and find that the AC-CPC objective significantly helps to improve the agent performance. We also show that data augmentation and future actions conditioning play an important role in the learning of representations to complexify the AC-CPC objective and help the model to make accurate future predictions. Following our early findings, we hope that this work will inspire researchers to further study the benefits of self-supervised learning techniques for model-based reinforcement learning.

\section{Acknowledgments}

This work was partly supported by The Alexander von Humboldt Foundation (AvH).

\bibliography{iclr2025_conference}
\bibliographystyle{iclr2025_conference}

\newpage

\appendix
\section{Appendix}

\subsection{Atari 100k Evaluation Curves}

\begin{figure}[!ht]
        \centering
        \includegraphics[width=0.9\linewidth]{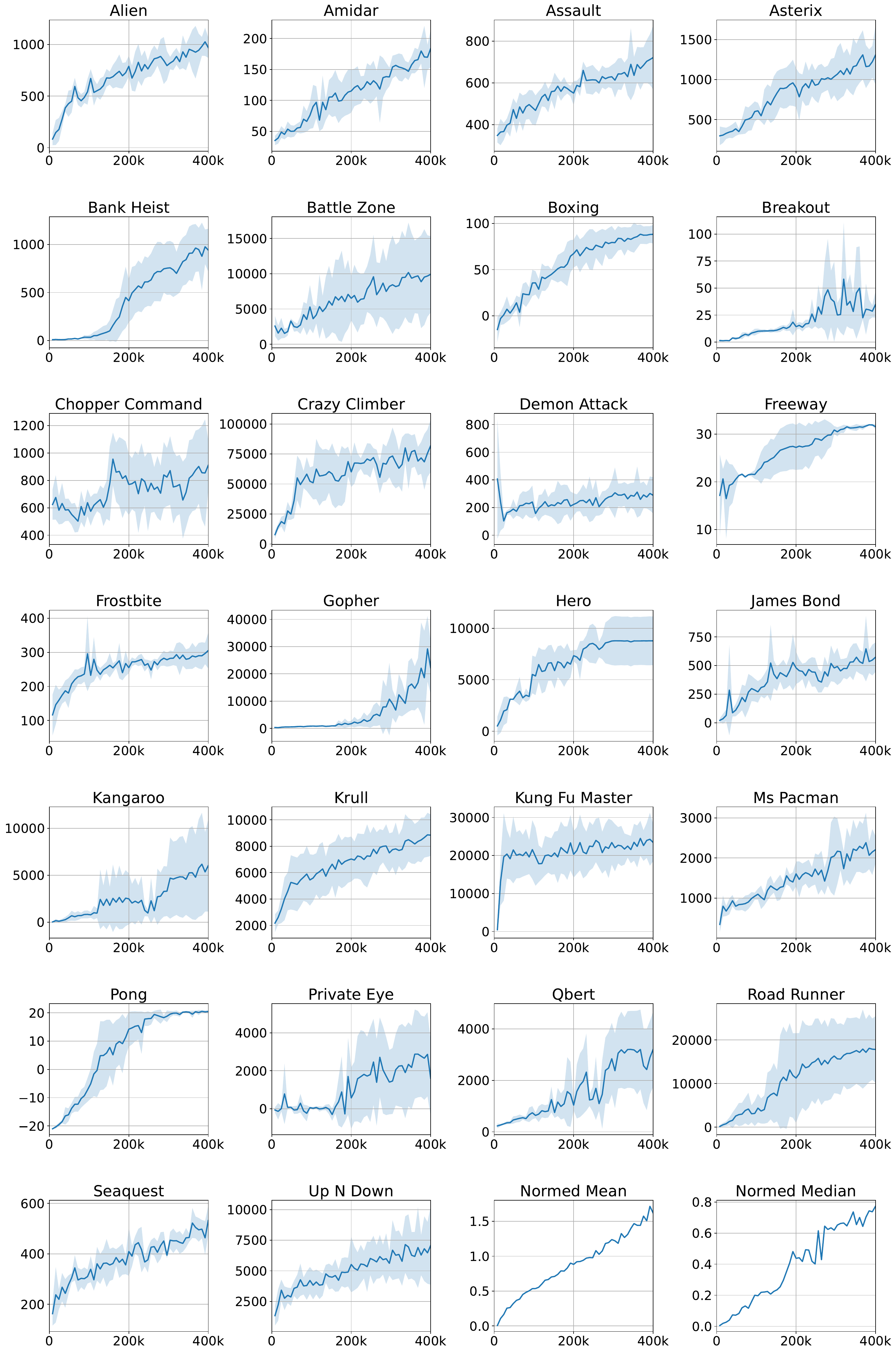}
        \caption{
        Evaluation curves of TWISTER on the Atari100k benchmark for individual games (400K environment steps). The solid lines represent the average scores over 5 seeds, and the filled areas indicate the standard deviation across these 5 seeds.}
        \label{appendix:results_atari100k}
\end{figure}

\newpage

\subsection{Model Architecture}

\begin{table}[!ht]
\caption{Architecture of the encoder network. The size of submodules is omitted and can be derived from output shapes. Each convolution layer (Conv) is followed by a layer normalization (LN) and a SiLU activation layer. The encoder downsamples images with strided convolutions layers using a kernel size of 4, a stride of 2 and a padding of 1. We flatten output features and project them to categorical distribution logits using a Linear layer. Stochastic states $z_{t}$ are sampled from Softmax probabilities and encoded to one hot vectors.}
\scriptsize
\centering
\hfill \break
\begin{tabular}{cc}
\toprule
Submodule & Output shape \\
\midrule
Input image ($o_{t}$) & $3 \times 64 \times 64$ \\
Conv + LN + SiLU & $32 \times 32 \times 32$ \\
Conv + LN + SiLU & $64 \times 16 \times 16$ \\
Conv + LN + SiLU & $128 \times 8 \times 8$ \\
Conv + LN + SiLU & $256 \times 4 \times 4$ \\
Flatten & $4096$ \\
Linear & $1024$ \\
Reshape + Softmax & $32 \times 32$ \\
Sample + One Hot (outputs $z_{t}$) & $32 \times 32$ \\
\bottomrule
\end{tabular}
\label{table:arch_encoder}
\end{table}

\begin{table}[!ht]
\caption{Architecture of the decoder network. Images are reconstructed from stochastic states. Each transposed convolution layer (ConvTrans) uses a kernel size of 4, a stride of 2 and padding of 1.}
\scriptsize
\centering
\hfill \break
\begin{tabular}{cc}
\toprule
Submodule & Output shape \\
\midrule
Input stochastic state ($z_{t}$) & $32 \times 32$ \\
Flatten & $1024$ \\
Linear & $4096$ \\
Reshape & $256 \times 4 \times 4$ \\
ConvTrans + LN + SiLU & $128 \times 8 \times 8$ \\
ConvTrans + LN + SiLU & $64 \times 16 \times 16$ \\
ConvTrans + LN + SiLU & $32 \times 32 \times 32$ \\
ConvTrans (outputs $\hat{o}_{t}$) & $3 \times 64 \times 64$ \\
\bottomrule
\end{tabular}
\label{table:arch_decoder}
\end{table}

\begin{table}[!ht]
\caption{Transformer block. Dropout~\citep{srivastava2014dropout} is used in each Transformer submodule to reduce overfitting. We also apply Dropout to attention weights in the MHSA module.}
\scriptsize
\centering
\hfill \break
\begin{tabular}{ccc}
\toprule
Submodule & Module alias & Output shape \\
\midrule
Input features (label as $x_{1}$) &  \multirow{5}{*}{MHSA} & \multirow{5}{*}{$T \times 512$} \\
Multi-head self-attention & & \\
Linear + Dropout & & \\
Residual (add $x_{1}$) & & \\
LN (label as $x_{2}$) & & \\
\midrule
Linear + ReLU & \multicolumn{1}{c}{\multirow{4}{*}{\begin{tabular}[c]{@{}c@{}}Feed \\ Forward\end{tabular}}} & $T \times 1024$ \\
Linear + Dropout & & $T \times 512$ \\
Residual (add $x_{2}$) & & $T \times 512$ \\
LN & & $T \times 512$ \\
\bottomrule
\end{tabular}
\label{table:arch_trans_block}
\end{table}

\begin{table}[!ht]
\caption{Transformer network. The stochastic states $z_{0:T-1}$ and one-hot encoded actions $a_{0:T-1} \in \mathbb{R}^{T \times A}$ are combined using an action mixer network~\citep{zhang2024storm}. The features are processed by the Transformer network to compute hidden states $h_{1:T}$.}
\scriptsize
\centering
\hfill \break
\begin{tabular}{ccc}
\toprule
Submodule & Module alias & Output shape \\
\midrule
Inputs stochastic states ($z_{0:T-1}$) & \multirow{5}{*}{Action Mixer} & $T \times 32 \times 32$\\
Flatten & & $T \times 1024$\\
Concat actions $a_{0:T-1}$ & & $T \times (1024 + A)$\\
Linear + LN + SiLU & & $T \times 512$\\
Linear + LN  & & $T \times 512$\\
\midrule
Transformer block $\times$ K & \multicolumn{1}{c}{\multirow{2}{*}{\begin{tabular}[c]{@{}c@{}}Transformer \\ Network\end{tabular}}} & \multirow{2}{*}{$T \times 512$} \\
Outputs hidden states ($h_{1:T}$) \\
\bottomrule
\end{tabular}
\label{table:arch_trans_network}
\end{table}

\begin{table*}[!ht]
\caption{Networks with Multi Layer Perceptron (MLP) structure. Inputs are first flattened and concatenated along the feature dimension. Each MLP layer is followed by a layer normalization and SiLU activation except for the last layer which outputs distribution logits.}
\scriptsize
\setlength{\tabcolsep}{6pt}
\centering
\hfill \break
\begin{tabular}{lccccc}
\toprule
Network & MLP layers & Inputs & Hidden dimension & Output dimension & Output Distribution \\
\midrule
Reward predictor & 3 & $s_{t}$ & 512 & 255 & Symlog Discrete \\
Continue predictor & 3 & $s_{t}$ & 512 & 1 & Bernoulli \\
Representation network & 2 & $z'_{t+k}$ & 512 & 512 & N/A \\
AC-CPC predictor & 2 & $s_{t}, a_{t:t+k}$ & 512 & 512 & N/A \\
Critic network & 3 & $s_{t}$ & 512 & 255 & Symlog Discrete \\
Actor network & 3 & $s_{t}$ & 512 & A & One hot Categorical\\
\bottomrule
\end{tabular}
\label{table:mlps}
\end{table*}

\newpage

\subsection{Hyper-parameters}

\begin{table*}[!ht]
\caption{TWISTER hyper-parameters. We apply the same hyper-parameters to all Atari games.}
\scriptsize
\setlength{\tabcolsep}{20pt}
\renewcommand{\arraystretch}{1.25}
\centering
\hfill \break
\begin{tabular}{lcc}
\toprule
Parameter & Symbol & Setting\\ 
\midrule
General & &\\
Batch Size & B & 16 \\
Sequence Length & T & 64 \\
Optimizer & --- & Adam~\citep{kingma2014adam} \\
Image Resolution & --- & $64 \times 64$ (RGB) \\
Training Step per Policy Step  & --- & 1 \\
Environment Instances & --- & 1 \\
\midrule
Transformer Network & & \\
Transformer Blocks & N & 4 \\
Number of Attention Heads & --- & 8 \\
Dropout Probability & --- & 0.1 \\
Attention Context  Length & --- & 8 \\
\midrule
World Model & & \\
Stochastic State Features & --- & 32 \\
Classes per Feature & --- & 32 \\
Dynamics Loss Scale & $\beta_{dyn}$ & 0.5 \\
Representation Loss Scale & $\beta_{reg}$ & 0.1 \\
AC-CPC Steps & K & 10 \\
Random Crop \& Resize Scale  & --- & (0.25, 1.0) \\
Random Crop \& Resize Ratio & --- & (0.75, 1.33) \\
Learning Rate & $\alpha$ & $10^{-4}$ \\
Adam Betas & $\beta_{1}$, $\beta_{2}$ & 0.9, 0.999 \\
Adam Epsilon & $\epsilon$ & $10^{-8}$ \\
Gradient Clipping & --- & 1000 \\
\midrule
Actor Critic & & \\
Imagination Horizon & H & 15 \\
Return Discount & $\gamma$ & 0.997 \\
Return Lambda & $\lambda$ & 0.95 \\
Critic EMA Decay & --- & 0.98 \\
% Critic EMA regularizer & --- & 1.0 \\
Return Normalization Momentum & --- & 0.99 \\
Actor Entropy Scale & $\eta$ & $3 \cdot 10^{-4}$ \\
Learning Rate & $\alpha$ & $3 \cdot 10^{-5}$ \\
Adam Betas & $\beta_{1}$, $\beta_{2}$ & 0.9, 0.999 \\
Adam Epsilon & $\epsilon$ & $10^{-5}$ \\
Gradient Clipping & --- & 100 \\
\bottomrule
\end{tabular}
\label{table:hyperparams}
\end{table*}

\newpage

\subsection{AC-CPC Predictions}
\label{appendix:cpc_preds}

\begin{figure}[!ht]
        \centering
        \includegraphics[width=0.9\linewidth]{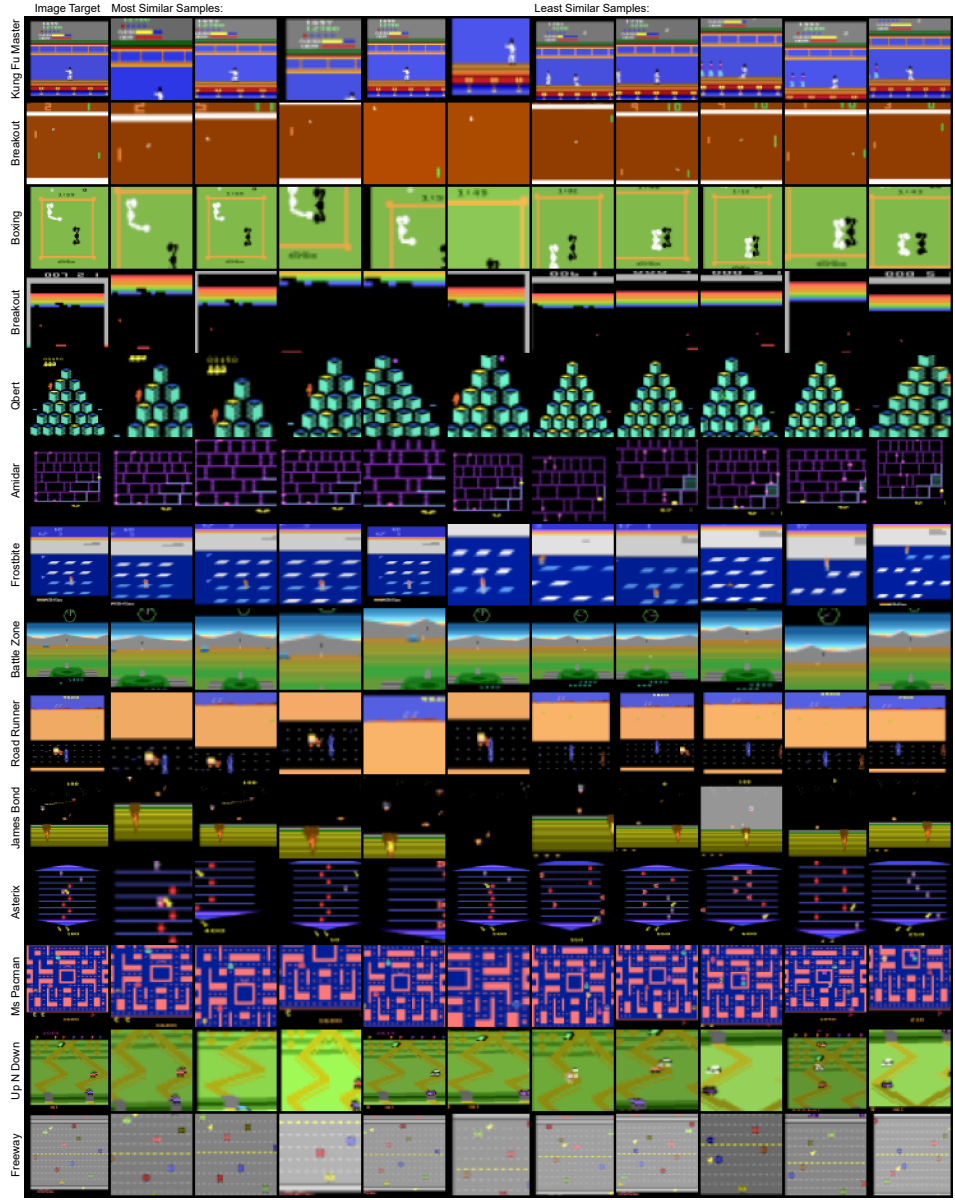}
        \caption{AC-CPC predictions made by the world model for diverse Atari games. We show the target positive sample without augmentation and predicted most/least similar samples among the batch of augmented image views. We observe that TWISTER successfully learns to identify most/least similar samples to the future target state using observation details such as the ball position in \textit{Pong}, the game score in \textit{Kung Fu Master} or the agent movements in \textit{Boxing}. AC-CPC necessitates the agent to focus on observation details to accurately predict future samples, thereby preventing common failure cases where small objects are ignored by the reconstruction loss.}
        \label{figure:cpc_preds}
\end{figure}

\newpage

\subsection{World Model Predictions}
\label{appendix:trajs_all}

\begin{figure}[!ht]
    \centering
    \includegraphics[width=0.9\linewidth]{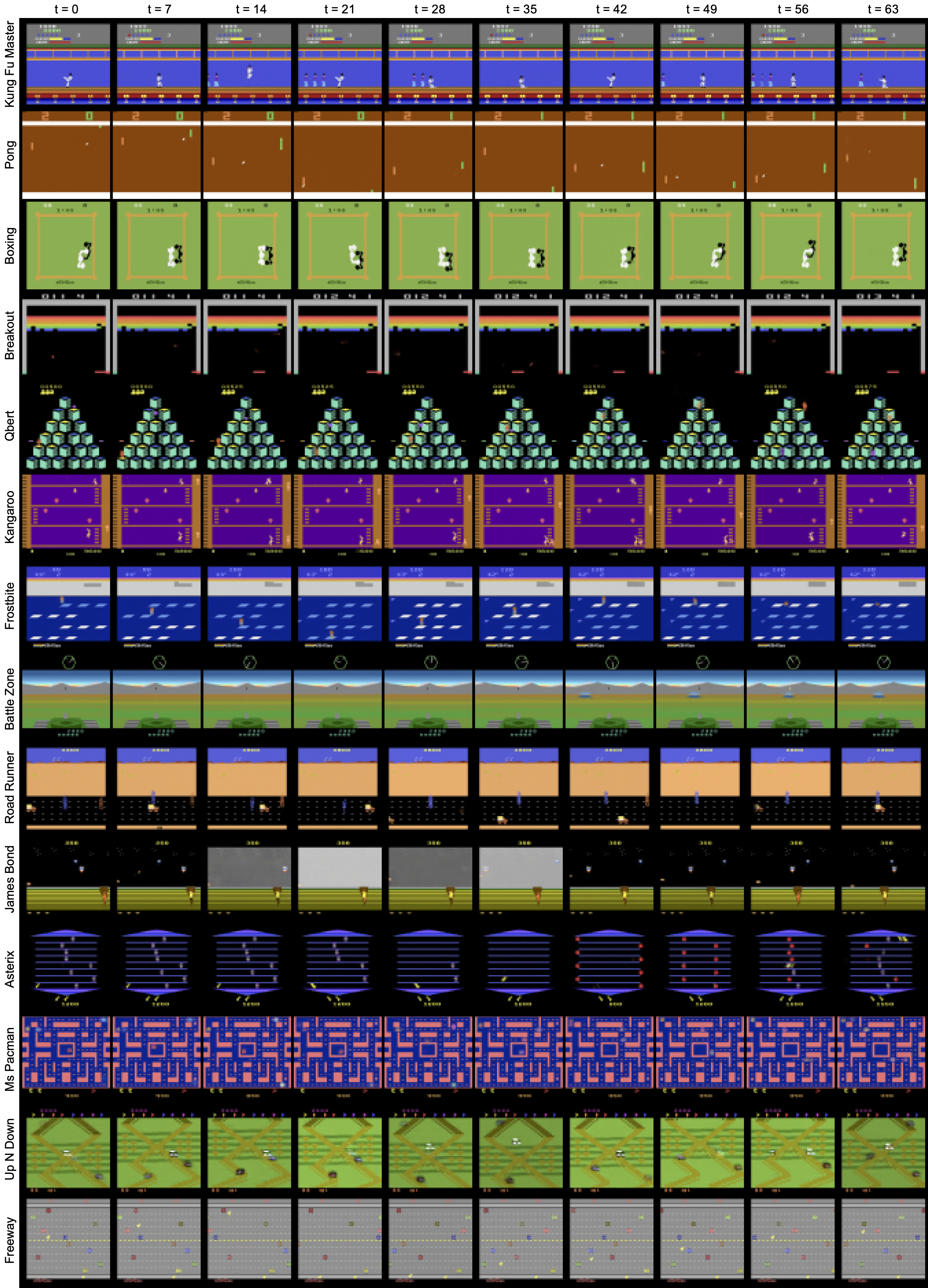}
    \caption{World Model Predictions. We show the decoder reconstruction of trajectories imagined by the world model over 64 time steps. We use 5 context frames and generate trajectories of 59 steps into the future using the Transformer network and dynamics predictor head. Actions are predicted by the actor network by sampling from the categorical distribution.}
    \label{figure:trajs_all}
\end{figure}

\newpage

\subsection{Ablations Results}
\label{appendix:main_ablations}

\begin{table}[!ht]
    \caption{Ablations of the AC-CPC loss, contrastive samples augmentation, conditioning on future actions and using DreamerV3's RSSM. We show agent scores and human-normalized metrics on the 26 games of the Atari 100k benchmark. The results are averaged over 5 seeds and bold numbers indicate best performing agent for each game.}
    \centering
    \setlength{\tabcolsep}{6pt}
    \scriptsize
    \hfill \break
    \begin{tabular}{lrrrrr}
    \toprule
    Game & TWISTER & No CPC & DreamerV3 World Model & No Action Conditioning & No Data Augmentation  \\ 
    \midrule
    Alien & 970 & 1147 & 1040 & 1101 & \textbf{1154}  \\
    Amidar & 184 & 173 & 174 & \textbf{191} & 154  \\
    Assault & 721 & \textbf{1168} & 735 & 711 & 657   \\
    Asterix & 1306 & 1165 & \textbf{1401} & 1082 & 1173   \\
    Bank Heist & 942 & 758 & \textbf{973} & 651 & 944   \\
    Battle Zone & 9920 & 5800 & \textbf{15540} & 12860 & 8980  \\
    Boxing & \textbf{88} & 81 & 82 & 77 & 84  \\
    Breakout & 35 & 14 & 34 & \textbf{59} & 31  \\
    Chopper Command & 910 & 984 & \textbf{1242} & 620 & 646   \\
    Crazy Climber & 81880 & \textbf{90680} & 89888 & 87272 & 77454  \\
    Demon Attack & 289 & 215 & \textbf{456} & 339 & 356  \\
    Freeway & \textbf{32} & \textbf{32} & 0 & 31 & 26  \\
    Frostbite & 305 & 714 & 571 & 884 & \textbf{953}   \\
    Gopher & \textbf{22234} & 1387 & 3318 & 2972 & 5851  \\
    Hero & 8773 & 8772 & 9944 & 7649 & \textbf{11079}  \\
    James Bond & \textbf{573} & 493 & 432 & 335 & 316  \\
    Kangaroo & \textbf{6016} & 4724 & 3816 & 1268 & 2668  \\
    Krull & 8839 & 8096 & 7469 & 8054 & \textbf{9065}  \\
    Kung Fu Master & 23442 & 22232 & \textbf{25518} & 19412 & 17566   \\
    Ms Pacman & 2206 & 2025 & 1691 & 1927 & \textbf{2294}  \\
    Pong & \textbf{20} & 13 & 18 & \textbf{20} & \textbf{20}  \\
    Private Eye & \textbf{1608} & 941 & 535 & 106 & 489  \\ 
    Qbert & 3197 & 2579 & 3542 & \textbf{4443} & 4231  \\
    Road Runner & \textbf{17832} & 10556 & 12254 & 12590 & 13348  \\ 
    Seaquest & 532 & 474 & \textbf{569} & 491 & 467   \\
    Up N Down & 7068 & 5816 & \textbf{30135} & 5378 & 7213   \\
    \midrule
    \# Superhuman & \textbf{12} & 8 & 11 & 9 & 8   \\
    Normed Mean (\%) & \textbf{\resultmean} & 112 & 121 & 111 & 120  \\
    Normed Median (\%) & \textbf{\resultmedian} & 44 & 69 & 42 & 68   \\
    \bottomrule
    \end{tabular}
    \label{table:results_atari100k_ablations}
\end{table}

\newpage

\subsection{DeepMind Control Suite Results}

We assess TWISTER's performance on continuous action control tasks by evaluating on the DeepMind Control Suite~\citep{tassa2018deepmind}. The suite was designed to serve as a reliable performance benchmark for reinforcement learning agents in continuous action space, including diverse control tasks with various complexities. Similarly to DreamerV3~\citep{hafner2023mastering}, we evaluate on 20 tasks using only high-dimensional image observations as inputs and a budget of 1M environment steps for training. 

We compare TWISTER with DreamerV3 and two other recent model-based approaches applied to continuous control. DreamerPro~\citep{deng2022dreamerpro} proposed a reconstruction-free variant of the Dreamer algorithm. Similarly to SwAV~\cite{caron2020unsupervised}, the agent learns hidden representations by encouraging consistent cluster assignments for different augmentations of the same images~\cite{caron2020unsupervised}. More recently, TD-MPC2~\cite{hansen2023td} extended the TD-MPC~\citet{hansen2022temporal} agent to multitask learning and demonstrated state-of-the-art performance on diverse continuous control tasks. TD-MPC unrolls its world model over the batch of sampled trajectories to predict the sequence of future latent states and environment quantities. The agent also learns a Q-value function to estimate long-term returns using Temporal Difference (TD) learning. It uses Model Predictive Control (MPC) for planning, selecting actions that maximize expected returns using world model value predictions.

Table~\ref{table:results_dmc_1M} shows the results obtained on the 20 tasks after 1M environment steps. We obtain DreamerPro\footnote{\texttt{https://github.com/fdeng18/dreamer-pro}} and TD-MPC2\footnote{\texttt{https://github.com/nicklashansen/tdmpc2}} results using official implementations. TWISTER obtains state-of-the-art performance with a mean score of 801.8. We also experiment with removing the AC-CPC objective and find that it has a positive impact on most of the tasks. AC-CPC particularly improves performance on complex tasks such as \textit{Acrobot Swingup}, \textit{Quadruped Run / Walk} and \textit{Walker Run}.

\begin{table*}[ht!]
    \caption{Agent scores on the DeepMind Control Suite under visual inputs. We show average scores over 5 seeds (1M environment steps). Bold numbers indicate best performing method for each task. We also underline TWISTER numbers to indicate tasks where AC-CPC improves performance.} 
    \setlength{\tabcolsep}{10pt}
    \scriptsize
    \centering
    \begin{tabular}{lccccc}
        \toprule
        Task & DreamerPro & DreamerV3 & TD-MPC2 & TWISTER (No CPC) & TWISTER (ours) \\ 
        \midrule
        Acrobot Swingup & \textbf{438.3} & 210.0   & 216.0 & 81.6 & \underline{239.4} \\
        Ball In Cup Catch & 962.2 & 957.1   & 717.2 & 964.8 & \textbf{\underline{966.8}} \\
        Cartpole Balance & \textbf{998.5} & 996.4   & 931.1 & \textbf{998.5} & 997.9 \\
        Cartpole Balance Sparse & \textbf{1000.0} & \textbf{1000.0}  & \textbf{1000.0} & \textbf{1000.0} & \textbf{1000.0} \\
        Cartpole Swingup & \textbf{871.0} & 819.1  & 808.1 & 759.1 & \underline{819.2} \\
        Cartpole Swingup Sparse & \textbf{811.2} & 792.9  & 739.0 & 524.4 & \underline{735.0} \\
        Cheetah Run & \textbf{898.3} & 728.7  & 550.3 & 695.1 & 694.0 \\
        Finger Spin & 600.6 & 818.5   & \textbf{986.0} & 845.8 & \underline{976.3} \\
        Finger Turn Easy & 879.6 & 787.7   & 788.9 & \textbf{956.4} & 923.7 \\
        Finger Turn Hard & 719.6 & 810.8   & 871.8 & \textbf{960.3} & 910.3 \\
        Hopper Hop & 252.6 & \textbf{369.6}   & 211.1 & 313.8 & \underline{313.9} \\
        Hopper Stand & 927.5 & 900.6  & 803.0 & \textbf{936.5} & 932.0 \\
        Pendulum Swingup & \textbf{845.2} & 806.3   & 743.2 & 529.2 & \underline{832.1} \\
        Quadruped Run & 616.2 & 352.3   & 361.9 & 503.8 & \textbf{\underline{652.1}} \\
        Quadruped Walk & 676.9 & 352.6   & 252.6 & 741.6 & \textbf{\underline{904.9}} \\
        Reacher Easy & 945.5 & 898.9  & \textbf{971.0} & 886.1 & \underline{933.1} \\
        Reacher Hard & 294.0 & 499.2   & \textbf{876.9} & 380.0 & \underline{565.8} \\
        Walker Run  & 750.0 & \textbf{757.8}   & 728.1 & 566.2 & \underline{711.2} \\
        Walker Stand & 974.8 & 976.7  & 915.8 & 969.5 & \textbf{\underline{976.9}} \\
        Walker Walk & \textbf{956.6} & 955.8   & 945.1 & 961.2 & 951.3 \\
        \midrule
        Mean & 770.9 & 739.6 & 720.9 & 728.7 & \underline{\textbf{801.8}} \\
        Median & 858.1 & 808.5 & 795.9 & 802.4 & \underline{\textbf{907.6}} \\
        \bottomrule
    \end{tabular}
    \label{table:results_dmc_1M}
\end{table*}

\newpage

\begin{figure}[!ht]
        \centering
        \includegraphics[width=0.9\linewidth]{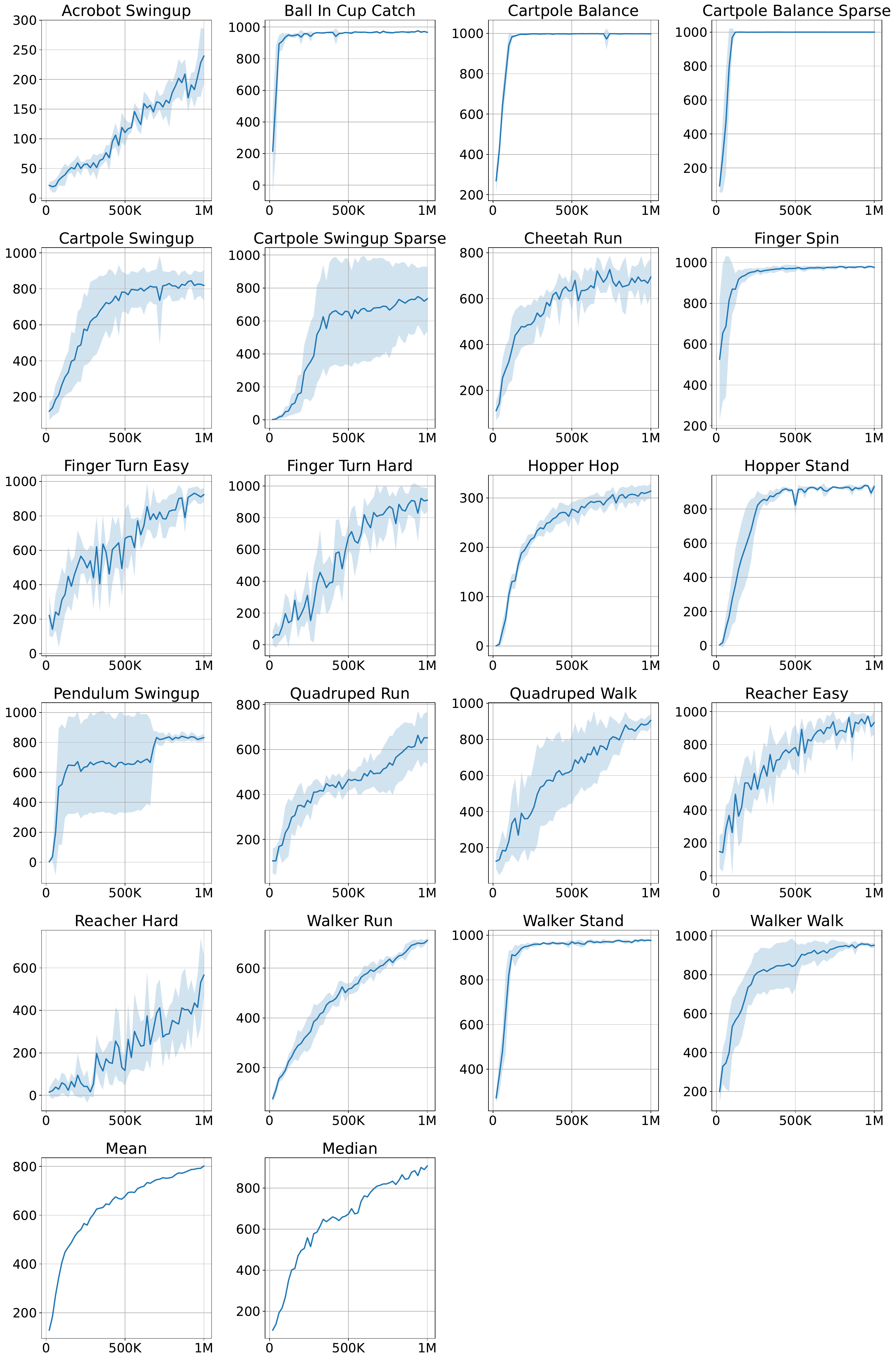}
        \caption{Evaluation curves of TWISTER on the DeepMind Control Suite for individual tasks (1M environment steps). The solid lines represent the average scores over 5 seeds, and the filled areas indicate the standard deviation across these 5 seeds.}
\end{figure}

\end{document}